\documentclass{article}
\usepackage{xcolor}
\usepackage{graphicx}
\usepackage{subcaption}
\usepackage{amsmath}
\usepackage{amssymb}
\usepackage{todonotes}
\usepackage{csquotes}

\usepackage[final]{corl_2020} 

\title{\enquote{What, not how} -  Solving an under-actuated insertion task from scratch}

\author{
  Giulia Vezzani\\
  \texttt{giulia.vezzani@google.com} \AND Michael Neunert \And Markus Wulfmeier \And Rae Jeong \And Thomas Lampe \And
Noah Siegel \And Roland Hafner \And Abbas Abdolmaleki \And Martin Riedmiller \And Francesco Nori}

\begin{document}
\maketitle

\begin{center}
DeepMind, United Kingdom
\end{center}


\begin{abstract}
Robot manipulation requires a complex set of skills
that need to be carefully combined and coordinated to solve a task. Yet, most  Reinforcement Learning (RL) approaches in robotics study tasks which actually consist only of a single manipulation skill, such as grasping an object or inserting a pre-grasped object. 
As a result the skill (`how' to solve the task) but not the actual goal of a complete manipulation (`what' to solve) is specified. In contrast, we study a complex manipulation goal that requires an agent to learn and combine diverse manipulation skills. We propose a challenging, highly under-actuated peg-in-hole task with a free,  rotational  asymmetrical peg, requiring a broad range of manipulation skills. While correct peg (re-)orientation is a requirement for successful insertion, there is no reward associated with it. Hence an agent needs to understand this pre-condition and learn the skill to fulfill it. The final insertion reward is sparse, allowing freedom in the solution and leading to complex emerging behaviour not envisioned during the task design.
We tackle the problem in a multi-task RL framework using Scheduled Auxiliary Control (SAC-X) combined with  Regularized  Hierarchical  Policy Optimization (RHPO) which successfully solves the task in simulation and from scratch on a single robot where data is severely limited.

\end{abstract}



\section{Introduction}

In recent years, Reinforcement Learning (RL) has been successfully applied to various robotic manipulation tasks, including grasping~\cite{lampe2013acquiring, levine2018learning,kalashnikov2018scalable}, non-prehensile manipulation~\cite{lowrey2018reinforcement,zeng2018learning}, cube stacking~\cite{riedmiller2018learning,wulfmeier2019compositional, nair2018overcoming,jeong2019self,zhu2018reinforcement}, insertion~\cite{vecerik2019practical,neunert2020continuous,schoettler2019deep}, and in-hand object reorientation~\cite{andrychowicz2020learning, nagabandi2020deep}.
Although relevant and challenging, each work individually only involves a fraction of the skills required for interacting with objects in the real world. In such tasks, an individual skill like grasping is often considered the final goal, whereas it is only a single part of solving real world manipulation. Some approaches specifically address the study of object affordance~\cite{nguyen2016detecting}, but their applications rarely go beyond pick-and-place~\cite{zeng2018robotic,florence2019self, manuelli2019kpam}. Non-prehensile manipulation tasks provide a playground for planning and decision making experiments but they are limited in the goals that can be achieved by using simple actions such as pushing and poking~\cite{lowrey2018reinforcement, zeng2018learning,jeong2019modelling}. Cube stacking can be solved with simple grasps without addressing advanced object manipulation such as reorientation. In-hand object orientation work has yielded impressive results, but the reorientation is often done without further purpose. In insertion tasks~\cite{vecerik2019practical, neunert2020continuous, schoettler2019deep}, the object is usually pre-grasped such that the task can be solved by moving the end-effector but without manipulating (grasping or reorienting) the object itself~\cite{levine2016end,vecerik2019practical}.

\begin{figure}
    \centering
    \begin{subfigure}{.4\textwidth}
    \centering
          \includegraphics[width=.7\linewidth]{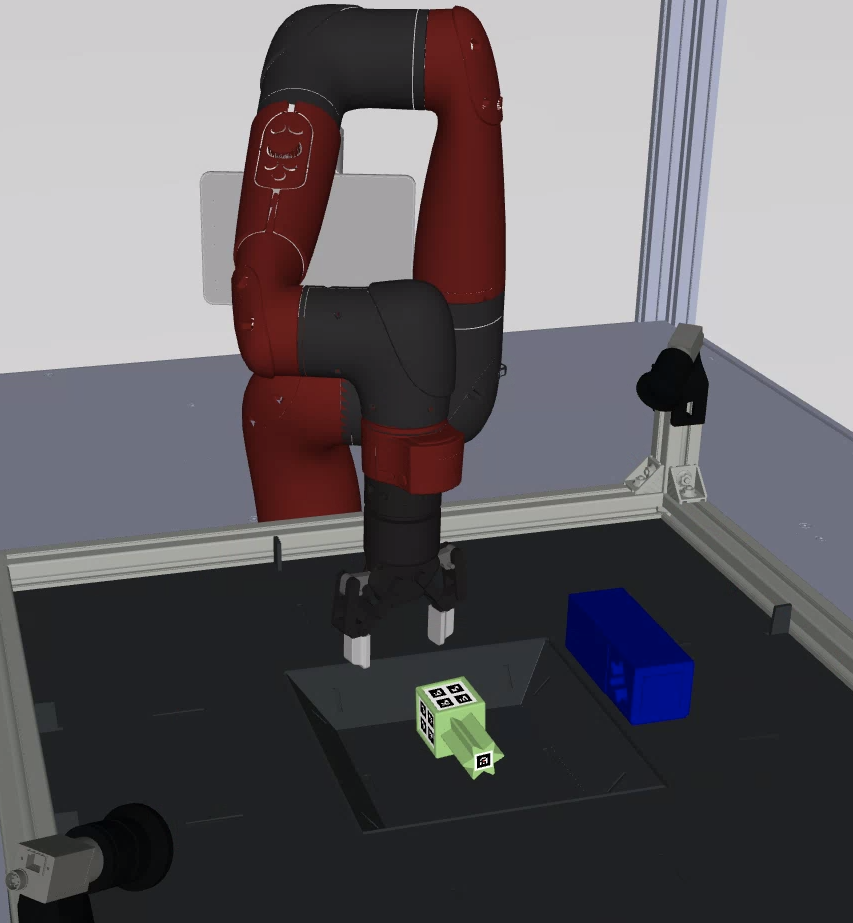}
          \caption{}
          \label{fig:sim_task}
    \end{subfigure}
    \begin{subfigure}{.4\textwidth}
    \centering
        \includegraphics[width=0.705\linewidth]{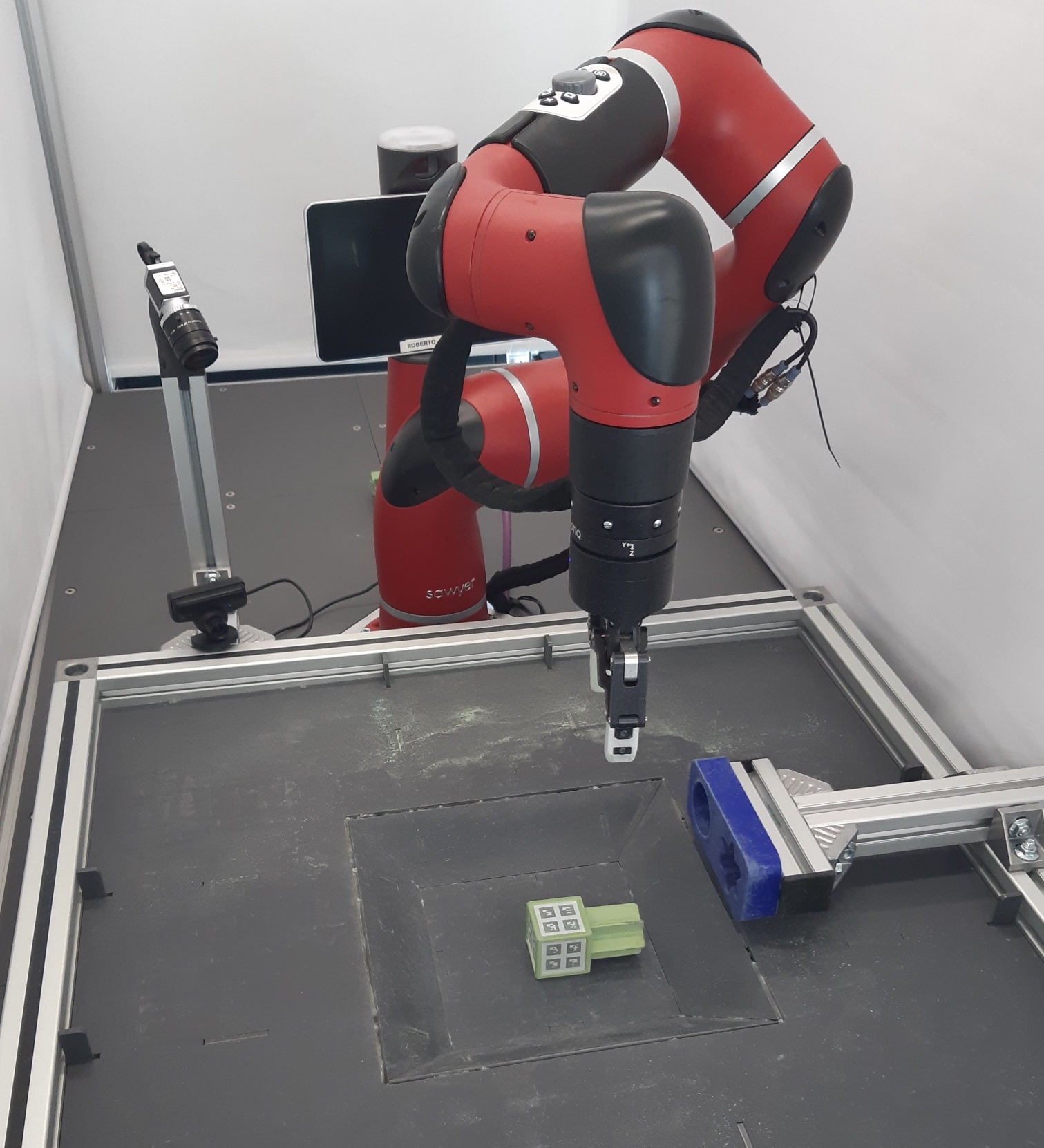}
        \caption{}
        \label{fig:real_task}
    \end{subfigure}
    \caption{Fig.~\ref{fig:sim_task} and \ref{fig:real_task} show respectively the simulated and real environment for the insertion task. The task consists of grasping the peg, reorienting (if needed) and inserting it inside a hole. The peg's tip and the hole have a star-shaped section for which only specific orientations make the insertion feasible.}
    \label{fig:task}
\end{figure}

In this work, we propose  and solve a novel peg-in-hole task that combines several forms of manipulation skills in a single task (Fig.~\ref{fig:task}). The task requires grasping a free, rotational asymmetrical peg, reorienting it (if needed), aligning it and final inserting it inside a rotational asymmetric hole. We explicitly design the task to pose the following challenges:

\begin{itemize}
    \item \textit{Rewarding success, not the solution path}. While the correct peg orientation is a requirement for successful insertion, there is no reward associated with it. Hence the agent needs to understand this pre-condition and how to reorient the peg to fulfill it, without receiving (intermediate) reward for it. Also, the task reward is sparse and only given for successful insertions. Hence, we specify \emph{what} the task is, but not \emph{how} to solve it. Thus, there is a range of viable strategies to solve the task, allowing for complex behaviours to emerge.
    \item \textit{Object reorientation with high under-actuation}. The robot's action space purposefully does not allow rotating the peg around the axis that is paramount to a successful insertion while the peg is grasped. To overcome this under-actuation, a rich manipulation behaviour -- potentially involving several re-grasps -- is required.
    \item \textit{Difficult exploration}. The task requires a complex behaviour involving several manipulation skills. Such a behaviour is highly unlikely to be discovered with random exploration. Additionally, it is hard to design a shaped reward for a (single task) RL algorithm.
    \item \textit{Credit assignment}. Inserting the peg into the hole with proper orientation requires executing a complex sequence of object manipulations, yet the insertion reward is sparse. Hence, an agent does not only need to discover a suitable sequence of motions but also correctly assign credit to any preceding manipulation.
\end{itemize}

We approach the problem using Scheduled Auxiliary Control (SAC-X)~\cite{riedmiller2018learning}, a multi-task RL framework. To further enhance data efficiency, we follow the example of Regularized Hierarchical Policy Optimization (RHPO)~\cite{wulfmeier2019compositional} where SAC-X is augmented with a hierarchical policy. Although SAC-X involves the definition of auxiliary policies to aid exploration, the task-completion reward is sparse in our case and we do not communicate the agent any reward concerning the peg reorientation required for solving the task. Therefore we do not provide crucial information on `how' to solve the task, but we focus instead on `what' the task is. This allows the final policy to develop original strategies for the reorientation and  insertion. To highlight the importance of a multi-task approach, we also provide a state-of-the-art single-task RL baseline using Expectation-Maximization based policy optimization algorithm (MPO~\cite{abdolmaleki2018maximum}) guided by a shaped reward.

Our experiments show that, despite  shaping the reward, regular RL algorithms struggle with exploration. 
In contrast, we show that a multi-task formulation such as SAC-X with RHPO can learn this complex insertion task from scratch, with impressive performance in simulation but also efficiently on a single real robot without human intervention.
Remarkably, the resulting complex emergent behaviours for reorienting the peg were not considered during the design of the task and the agent was not rewarded for them during the training. 
To the best of our knowledge, this is the first time such a complex manipulation task is learned from scratch on a real robot.

To summarize, the main contributions of this work are 1) the \textit{design} of a novel peg-in-hole task, which combines several manipulation skills; 2) its \textit{solution from scratch} not only in simulation but also on a single real robot without human intervention or significant reward shaping; 3)  the \textit{emergence} of smart under-actuated behaviours for reorienting the peg thanks to the fact that we reward the  success, not the solution path.


  
\section{Related work} 
\label{sec:related_work}

Many robotic manipulation tasks have been tackled with RL, spanning from grasping to in-hand object reorientation. Grasping has been one of the first real world applications where RL algorithms have provided impressive performance on real robots~\cite{lampe2013acquiring, levine2018learning,kalashnikov2018scalable}. 
Despite being a key skill for manipulating objects, grasping is often only a sub-step, especially in more complex manipulation tasks. Recent works have tackled more long-horizon manipulation objectives, such as stacking~\cite{riedmiller2018learning, wulfmeier2019compositional} and  non-prehensile manipulation tasks~\cite{lowrey2018reinforcement, zeng2018learning}. Agents exhibit impressive planning strategies~\cite{riedmiller2018learning, wulfmeier2019compositional} and utilization of the environments~\cite{lowrey2018reinforcement, zeng2018learning} in these experiments. But the resulting behaviors are still limited, as stands, for example, for cube grasping and object pushing. In the recent years, there has been an increasing interest in in-hand manipulation tasks such as object reorientation~\cite{andrychowicz2020learning, nagabandi2020deep, zhu2019dexterous, van2015learning, akkaya2019solving}. Because they are under-actuated, these tasks require a great level of dexterity and control skills but, again, only study a sub-problem of manipulation. In addition, exploration is heavily guided by shaped rewards imposing the risk of biasing the agent towards sub-optimal solutions~\cite{andrychowicz2020learning, nagabandi2020deep}. Another category of manipulation tasks solved on real robots with RL includes peg-in-hole insertions~\cite{vecerik2019practical, neunert2020continuous, schoettler2019deep}.  The main focus of these works is usually on the compliance and precision required for properly inserting the peg inside the hole. However, the peg is usually pre-grasped or fixed on the end-effector and robot's action space is such that the task is fully-actuated. Hence, the agent is not required to learn to grasp or reorient an object.

The task we consider in this work differs from the ones above mentioned because it combines together several challenges: 1) it requires several manipulation skills (peg grasping, reorientation and insertion) in order to solve the task, making exploration difficult; 2) the grasping strategy required for it is not trivial, calling for advanced object reorientation strategies to overcome the significant under-actuation of the robot (more on this in Section~\ref{sec:task}); 3) it consists of a complete manipulation task that demands a sequence of multiple object manipulations to be accomplished and, hence, an agent has not only to discover the suitable sequence but also correctly assign credit.

Real world manipulation tasks pose also the challenge of dealing with robotic hardware, which makes RL training difficult due to the time required for collecting the necessary data and the safety measurements needed for doing so.
Several works circumvent these issues by using demonstrations~\cite{rajeswaran2017learning, vecerik2019practical,zhu2019dexterous, cabi2019scaling}, transferring simulated solutions to the real world~\cite{jeong2019self, james2019sim, tobin2017domain} or by learning a model of the environment~\cite{nagabandi2020deep}. In this work instead we show that, similarly to~\cite{riedmiller2018learning, wulfmeier2019compositional,deisenroth2014multi}, we can efficiently learn a complex manipulation task by training multi-task RL agents from scratch using experience from only one single real robot.



	

\section{Background} 
\label{sec:background}
The goal of this paper is to analyze if state-of-the-art Reinforcement Learning algorithms can solve a complete manipulation task, consisting of grasping, reorienting and inserting a peg inside a hole.
 As it will be shown in the Section~\ref{sec:result}, we consider two state-of-the-art algorithms: Maximum a-posteriori Policy Optimization (MPO)~\cite{abdolmaleki2018maximum} and Scheduled Auxiliary Control (SAC-X)~\cite{riedmiller2018learning} combined with Regularized Hierarchical Policy Optimization (RHPO)~\cite{wulfmeier2019compositional}.
 
\subsection{Reinforcement Learning problem formulation}
We consider the problem of Reinforcement Learning (RL)
in a Markov Decision Process (MDP). Let $s \in \mathbb{R}^
S$ be the state of the
agent in the MDP,
$a \in \mathbb{R}^A$
 the action vector and $p(s_{t+1}|s_t, a_t)$ the probability
density of transitioning to state $s_{t+1}$ when executing action
$a_t$ in $s_t$. All actions are assumed to be sampled from a policy
distribution $\pi_{\theta}(a|s)$, with parameters $\theta$. After executing an
action – and transitioning in the environment – the agent
receives a scalar reward $r(s_t, a_t)$.
The goal of
Reinforcement Learning is defined as maximizing the sum of discounted expected rewards $E_{\pi}[R(\tau_{0:\infty})] = E_{\pi}
[\sum_{t=0}^{\infty}\gamma_t
r(s_t, a_t) |
a_t \sim \pi(\cdot|s_t), s_{t+1} \sim p(\cdot|s_t, a_t), s_0 \sim p(s)]$, where $p(s)$
denotes the initial state distribution or, if assuming random
restarts, the state visitation distribution, and $\tau_{t:\infty} = {(s_t, a_t), . . . }$ refers to the trajectory
starting in state $s_t$.

\subsection{Scheduled Auxiliary Control (SAC-X)}
One of the main challenges of RL is given by the kind of reward $r(s_t, a_t)$ provided by the environment. Sparse task-completion rewards are easy to define and avoid biasing the agent towards sub-optimal solution of the task. However, they pose an exploration challenge, since the agent does not receive any useful feedback on its actions until the task is completed.
SAC-X proposes a solution for exploration with sparse rewards by augmenting the standard RL problem with
a set of auxiliary intentions $I = \{I_1, \dots, I_k\}$. These intentions are modeled as MDPs  that share the state, observation and action space as well as the transition dynamics
with the main task $M$, but have separate auxiliary reward functions $r_{I_1}(s, a), \dots , r_{I_k}(s, a)$. This formulation assumes full control over the auxiliary rewards; i.e. the auxiliary reward can be computed and evaluated at any state action pair. There are several simple auxiliary rewards that can be easily obtained on a real robot to improve exploration. 
Given the set of reward functions we can define intention
policies and their sum of discounted rewards as $\pi_{\theta}(a|s,T)$ and
$E_{\pi_{\theta}}(a|s,T)[
R_{T} (\tau_{t:\infty})] = E_{\pi_{\theta}}(a|s,T)[\sum_{
t=0}^{\infty}\gamma_t r_{\tau} (s_t, a_t)]$,
where $T = I \cup {M}$.
SAC-X allows training  the auxiliary intentions policies together with the main task
policy to achieve their respective goals and, at the same time,  utilizes the
intentions for fast exploration in the main sparse-reward
task $M$. This is accomplished by using a hierarchical
objective for policy training consisting of  two parts.
The first part is given by a joint
policy improvement objective for all intentions. Through the definition of a proper joint policy improvement objective each intention is optimized to select the optimal action for its task.
The second part of the hierarchical objective instead is concerned with learning a scheduler that
sequences intention-policies. In this regard, a proper objective is defined by considering the period and the number of switches allowed to the scheduler and modeling the behaviour of the scheduler in the policies. When optimizing the scheduler, the individual intentions as considered fixed.  The optimization of the intentions and scheduler objectives requires some considerations typical of off-policy and multi-task settings that are outside the scope of this review. For more details on it, please refer to~\cite{riedmiller2018learning}.

\subsection{Regularized Hierarchical Policy Optimization (RHPO)}

Regularized Hierarchical Policy Optimization (RHPO) is an algorithm for robust training of
hierarchical policies in multitask settings. The policy structure proposed within the RHPO approach consists of a set of low-level components combined with a high-level controller. When training in multi-task settings, the low-level components are shared across tasks and shielded from information about any specific task. Task information is communicated only to the high-level controller, allowing the low-level components to learn task-independent skills. RHPO provides an efficient way to specifically optimize this hierarchical structure when learning for multiple tasks. It is built on an
Expectation-Maximization based policy optimization algorithm
(similar to MPO~\cite{abdolmaleki2018maximum}), which is adapted to the application to hierarchical policies in the multitask case.  RHPO  can efficiently optimize hierarchical
policies in a multitask setting thanks to a robust off-policy
learning schemes that uses all transition data to train
each low-level controller independent of the actually executed
one.

RHPO can be combined effectively with SAC-X, as already shown in the original work. In particular, RHPO provides a method to optimize a hierarchical policy in a multi-task setting, while SAC-X manages the multi-task learning itself through the definition of objectives for the intentions and the scheduler. 
More details are available in~\cite{wulfmeier2019compositional}.

\section{Description of the task}
\label{sec:task}
The task we consider in this paper consists of reorienting (if needed), grasping and inserting a peg inside a hole using an industrial robot arm equipped with a two-finger gripper. The action space is 5D, i.e. the agent only controls the Cartesian linear velocity of the end-effector, the rotational velocity of the gripper around the vertical axis and the gripper opening/closing speed. Therefore, only top grasps are possible.
Figs.~\ref{fig:sim_task} and~\ref{fig:real_task} show respectively the simulated environment and real setup.
In both cases, we use a Rethink Robotics Sawyer robot arm equipped with Robotiq gripper (for details on the setup, see Appendix \ref{appendix:robot} and \ref{appendix:sim}).

\begin{figure}
    \centering
    \begin{subfigure}{.3\textwidth}
    \centering
          \includegraphics[width=.81\linewidth]{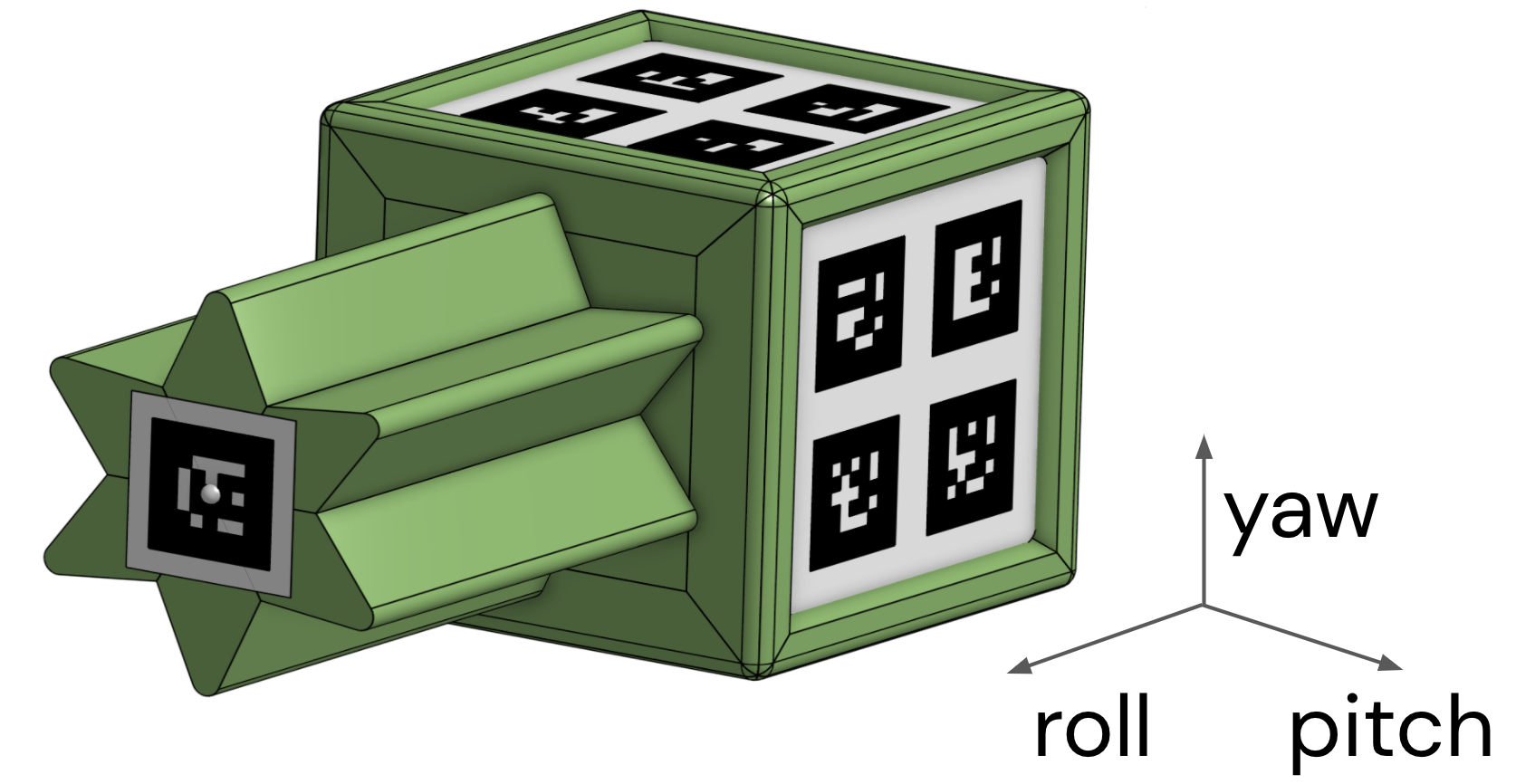}
          \caption{}
          \label{fig:peg}
    \end{subfigure}
    \begin{subfigure}{.3\textwidth}
    \centering
        \includegraphics[width=.8\linewidth]{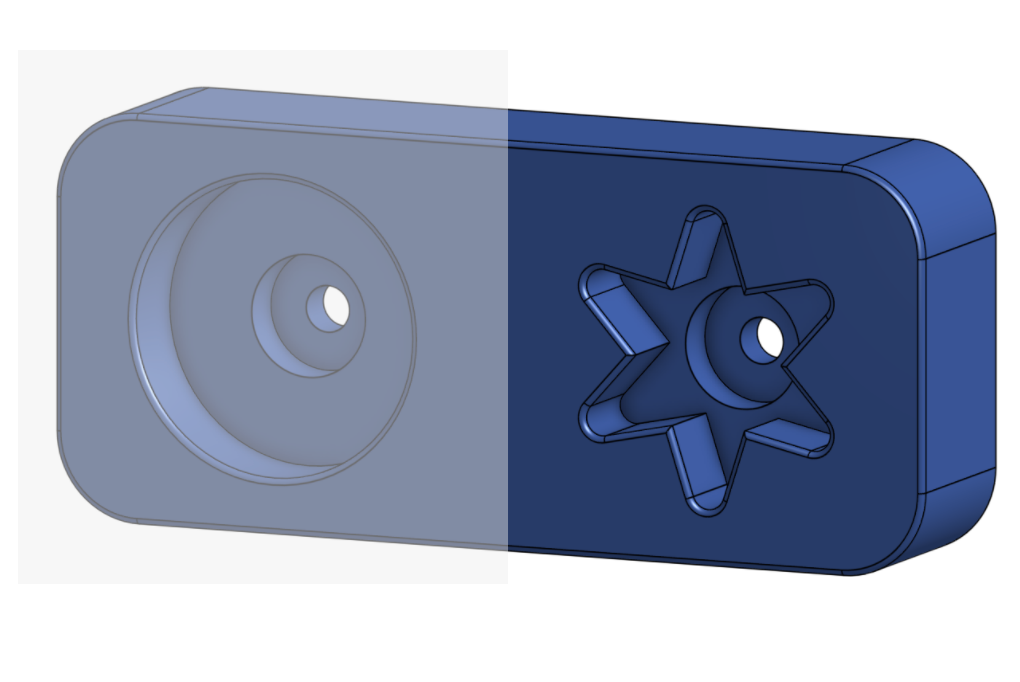}
        \caption{}
        \label{fig:hole}
    \end{subfigure}
    \hspace{10pt}
    \begin{subfigure}{.3\textwidth}
        \includegraphics[width=0.8\linewidth]{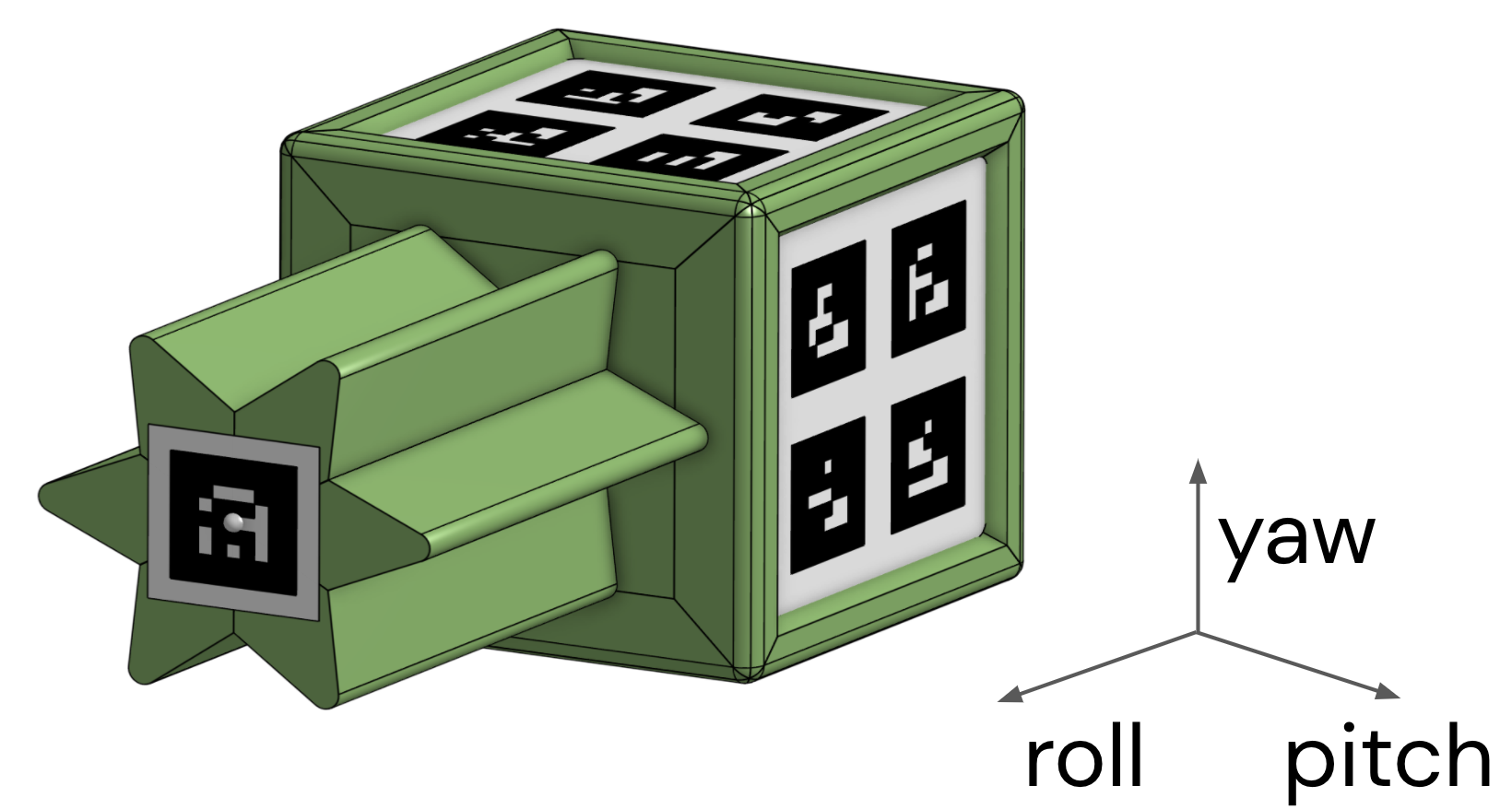}
        \caption{}
        \label{fig:wrong_orient}
    \end{subfigure}
    \caption{Peg~\ref{fig:peg} and hole~\ref{fig:hole} CAD models. The same models have been used in simulation and to 3D print the objects for the real setup. The blue socket includes also a circular hole, which has not been used during our experiments. Figs.~\ref{fig:peg} and ~\ref{fig:wrong_orient} show respectively the peg in the correct and wrong orientation for the insertion task.}
    \label{fig:objects}
\end{figure}
The peg features a cuboid handle and a six-corner star-shaped tip (Fig. \ref{fig:peg}). The hole has a matching six-corner star-shaped section (Fig. \ref{fig:hole}).
The peg design is such that, when dropped, the peg's roll axis is horizontal with two possible orientations of the tip (ignoring symmetries) (Fig.~\ref{fig:peg} and~\ref{fig:wrong_orient}). 
When the tip's initial orientation is the same as the one required for the insertion (Fig.~\ref{fig:peg}), the robot can grasp the peg, align the roll axis with the hole and insert it, without facing under-actuation. If the initial orientation of the tip is not matching the hole (Fig.~\ref{fig:wrong_orient}), the tip needs to be reoriented first, requiring a rotation around the roll axis of the peg. However, the Cartesian controller -- by design -- does not allow such an end-effector rotation. Hence, reorientation while holding the peg is impossible and the agent has to circumvent this under-actuation.

In summary, the object and setup design together with the robot action space challenges the agent to learn to reorient the peg in an under-actuated manner.

\section{Experimental results}
\label{sec:result}

Within this work, we use the insertion task to answer two research questions:
\begin{enumerate}
    \item Can RL solve such a complicated manipulation task from scratch in simulation and, most importantly, on a single robot where data collection is time consuming?
    \item Given that the agent is not provided with information on \textit{how} to solve the task, do any interesting behaviours emerge in order to deal with the challenges posed by the task design?
\end{enumerate}
The next two sections answer these questions taking advantage of the experiments we carried out, with a particular emphasis on those executed on the real robot.


\subsection{Task solution}
Among the challenges arising from the insertion task are \textit{difficult exploration} and \textit{credit assignment}. The agent has to combine several skills (e.g. grasping, lifting and orienting the peg) in a specific order to complete the insertion. Given the large state space and sparse reward provided within the task, we cannot expect to experience successful trials within a reasonable number of episodes using purely random exploration.

We evaluate two different approaches to deal with these challenges. For our baseline, we use Maximum a-posteriori Policy Optimization (MPO)~\cite{abdolmaleki2018maximum} (see Section~\ref{sec:mpo}), a state-of-the art RL algorithm. To avoid the sparse reward problem, we design a shaped reward as the sum over rewards for the individual skills. However, such a shaped reward turns out not to be enough for learning all skills and neither their proper execution order, due to poor exploration. Hence, in our main approach, we enforce a more structured exploration strategy by casting the insertion task into a multi-task problem (see Section~\ref{sec:multi_task}). The learning paradigm we use in this case is  Scheduled Auxiliary Control (SAC-X)~\cite{riedmiller2018learning} combined with  Regularized Hierarchical Policy Optimization (RHPO)~\cite{wulfmeier2019compositional}. 

\subsubsection{Baseline experiments with shaped reward}
\label{sec:mpo}
The shaped reward we design explicitly encourages the robot to solve the insertion plus some of the skills required:
\begin{equation}
     R = \frac{1}{N} \sum_{i=1}^N{R_{skills, i}},
     \label{eq:shaped_reward}
 \end{equation}
 with $N$ equal to the number of skills plus the insertion task.
 A detail description of the skill rewards is available in Appendix \ref{appendix:shaped_reward}.
It is worth highlighting that those rewards do not include -- by design -- any information on the peg's tip orientation required to properly insert it into the hole.  More specifically, none of the rewards $R_{skills, i}$ encourages the required rotation around the roll axis, i.e. they do not specify that the peg's tip orientation has to match the hole orientation. The agent has to infer this correlation by itself when attempting to insert the peg into the hole.
 
 We trained MPO with the shaped reward of Eq. \ref{eq:shaped_reward} both in simulation and on the real robot. Details on the training and hyperparameters used for this are collected in the Appendix \ref{appendix:mpo_params}.
Despite the amount of information encoded within the reward, the agent is not able to solve the task  neither in simulation nor on the robot. In particular, the agent  learns only a subset of the skills required for solving the insertion (specifically to reach and push the object close to the hole),  and eventually gets stuck in a local minimum, never achieving the maximum reward. More details on these results in Appendix~\ref{appendix:mpo}. While there might indeed be a shaped reward such that MPO could solve the task, it is hard to design and time-intense to test and verify. It might require several experiments, each lasting for several days or weeks, to tune it.

\subsubsection{Experiments with multi-task problem formulation}
\label{sec:multi_task}
In our multi-task approach, we guide the exploration by providing several auxiliary intentions, where each of them represents a skill potentially useful to solve the insertion. The main purpose of these intentions is to provide the agent with useful data for learning the task. Yet, solving the intentions is secondary with respect to the main task. If needed, the agent can therefore disregard some of the auxiliary skills altogether or develop new behaviors it is not rewarded for by an intention. For the auxiliary intentions we use the rewards $R_{skills,i}$ introduced in Eq.~\ref{eq:shaped_reward} and, therefore, none of the  intentions rewards the agent for reorienting the tip of the peg around the roll axis.
Consequently, although the auxiliary intentions do guide overall exploration, they do not specify the hardest part of the task (the peg reorientation around its roll axis), leaving it therefore to the agent to discover it is needed and \textit{how} to achieve it.  We specify instead \textit{what} the task is by using a sparse reward for the main insertion task. We train our policy using SAC-X combined with RHPO~\cite{wulfmeier2019compositional}. More information on the multi-task formulation is available in Appendix \ref{appendix:multi_task}.

\begin{figure}
    \centering
    \begin{subfigure}{.45\textwidth}
    \centering
          \includegraphics[width=.8\linewidth]{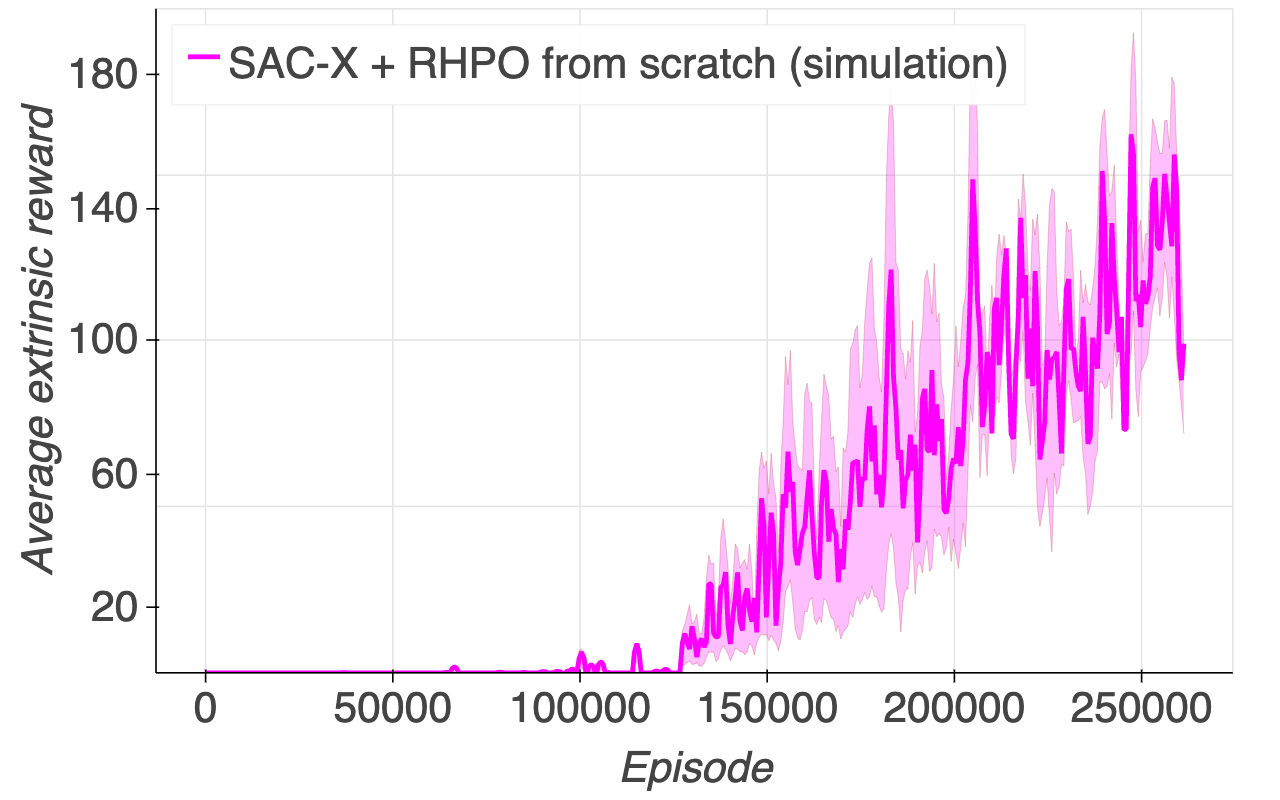}
          \caption{}
          \label{fig:sim_plot}
    \end{subfigure}
    \begin{subfigure}{.45\textwidth}
    \centering
        \includegraphics[width=.8\linewidth]{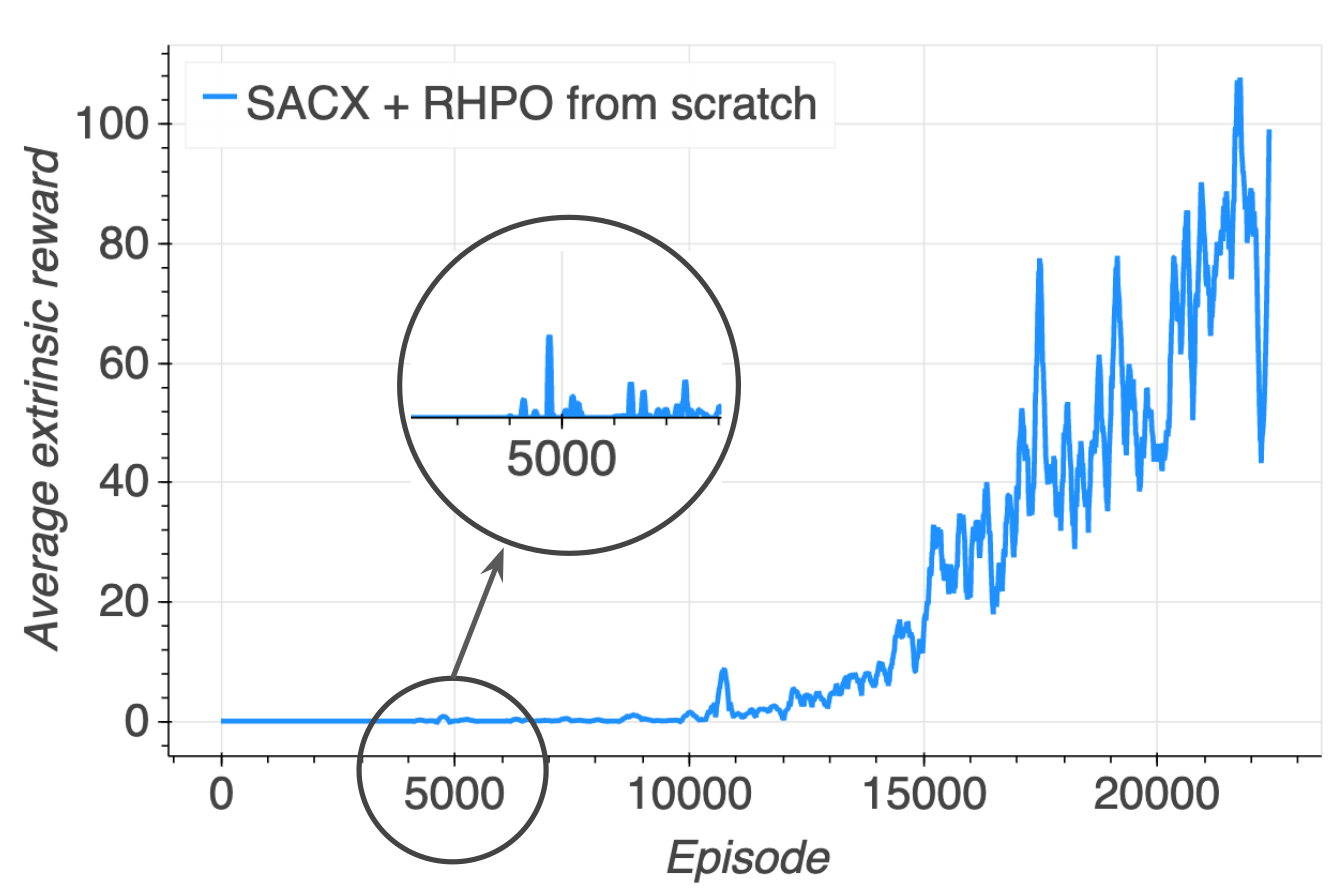}
        \caption{}
        \label{fig:real_plot}
    \end{subfigure}
    \caption{Averaged main task reward (window size: 50) in simulation (Fig.~\ref{fig:sim_plot}) and on the real robot (Fig~\ref{fig:real_plot}).
    Since the results in simulation have been obtained in a distributed fashion using 64 actors, a fair comparison between sim and real requires a detailed explanation, provided  in Appendix~\ref{sec:sim_vs_real}.
    Note that the theoretical maximum reward is 600, for an episode initialized with the peg already inserted. Higher rewards therefore represent faster or more reliable completions of the task.}
    \label{fig:sim_real_plots}
\end{figure}

The agent we obtained in this case learns to solve the task, both in simulation (Fig.~\ref{fig:sim_plot}) and on the real robot (Fig.~\ref{fig:real_plot}). In the latter case in particular, the agent manages to successfully insert the peg inside the hole for the first time after 5000 episodes (nearly 5 days of training). After 21000 episodes of training (nearly 21 days) the agent can solve the task with a high success rate of 0.85 when the peg's tip is initially in the correct orientation and 0.73 if the peg instead requires to be reoriented. The unsuccessful tries usually stem from the agent not being able to reorient the peg in time.
As it will be discussed in Section~\ref{sec:emergent_behaviours}, the reorientation strategy for the peg, due to the robot's under-actuation, depends partly on chance and may require multiple trials to succeed. But even when the tip is initially properly oriented, it can happen that the robot drops the peg during manipulation, which can flip it and then requires reorientation.

\begin{figure}
    \centering
    \begin{subfigure}{.45\textwidth}
    \centering
          \includegraphics[width=.8\linewidth]{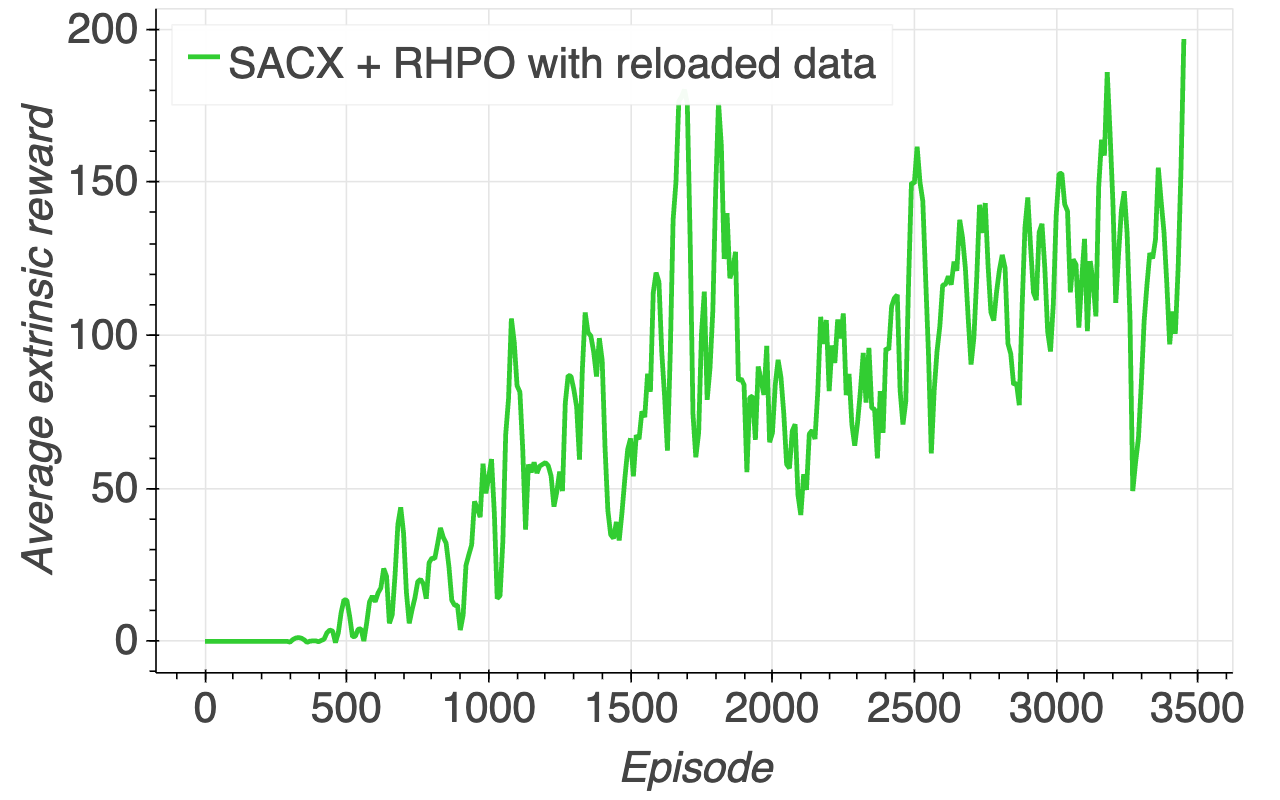}
          \caption{}
          \label{fig:reloaded_plot}
    \end{subfigure}
    \begin{subfigure}{.45\textwidth}
    \centering
        \includegraphics[width=.8\linewidth]{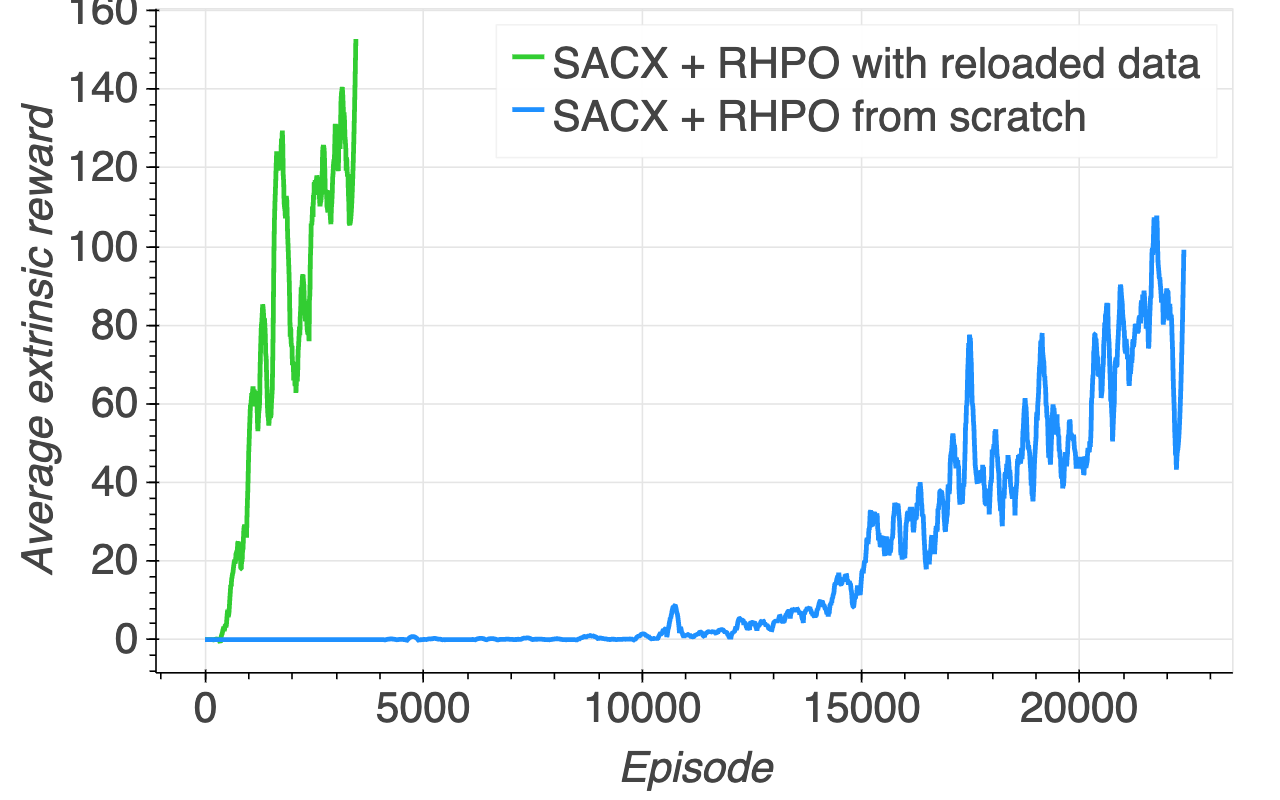}
        \caption{}
        \label{fig:real_vs_reloaded}
    \end{subfigure}
    \caption{Fig.~\ref{fig:reloaded_plot} shows the averaged  main task reward (window size: 50) when using SAC-X combined with RHPO and reloading the data collected during the previous training from scratch. After only 500 episodes (half a day of training on the robot), the agent is able to insert the peg inside the hole. After 3500 new episodes the agent achieves a higher reward than the one obtained when training from scratch (see Fig.~\ref{fig:real_vs_reloaded} for the comparison). This represents not only an increased success rate but also the ability of the agent to solve the task much faster within a single episode. }
    \label{fig:real_reloaded_plots}
\end{figure}

While the task can be successfully learned from scratch, we also execute a further training run (using the same training setup). We start with an untrained policy and critic but reload the data from the previous training run. The new agent then learns from this reloaded data as well as newly collected data added during the second training run.
 Providing the algorithm with data that partially contains successful insertions already at the beginning of training leads to even better performance within just a few days of running the new training on the real robot (first insertions happen after only 500 episodes, nearly half a day of training, see Fig.~\ref{fig:real_reloaded_plots}). The freshly trained policy is able to insert the peg inside the hole with 0.90 success rate, independent of whether the peg is initially in the correct orientation or not. In particular, the agent improves the reorientation strategy making it more robust and faster than in the previous experiment and reducing the cases when it does not reorient the peg in time (see Fig.~\ref{fig:real_vs_reloaded}). We provide examples of the execution of these two successful policies trained from scratch and with reloaded data in the supplementary video\footnote{\href{https://youtu.be/JnvdNpWAia4}{https://youtu.be/JnvdNpWAia4}}.

In conclusion, the experiments show how a multitask RL approach such as SAC-X coupled with RHPO, can solve a very challenging, complex manipulation task even from scratch on a single real robot.

\subsection{Emergent behaviours}
\label{sec:emergent_behaviours}
The second research question we aim to answer is whether smart behaviours emerge in the solution of the task.
For this analysis we focus on the multi-task experiments carried out on the real robot.

As mentioned in the previous sections, we do not specify the reorientation of the peg's tip as one of the skills to be learned. As a result, the agent does not only have to explore a reorientation strategy. Due to the sparse reward, it also needs to subsequently insert the peg successfully \emph{and} correctly assign credit to the reorientation maneuver.

Learning to properly and reliably reorient the peg is also challenging, since the agent needs to develop an under-actuated strategy to exploit the limited action space of the robot. In fact, if the peg's tip is initially in the wrong orientation (Fig.~\ref{fig:wrong_orient}), the robot cannot simply reorient the peg by grasping it and rotating the wrist, because that rotation is not enabled by our 5D controller.

\begin{figure}[t!]
    \centering
    \begin{subfigure}{.24\textwidth}
    \centering
          \includegraphics[width=1.\linewidth]{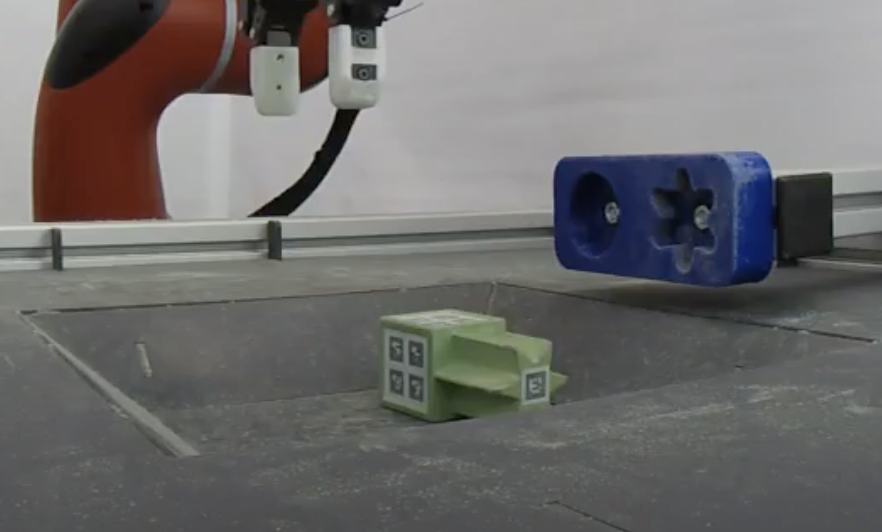}
          \caption{}
          \label{fig:s1}
    \end{subfigure}
    \begin{subfigure}{.24\textwidth}
    \centering
        \includegraphics[width=1.\linewidth]{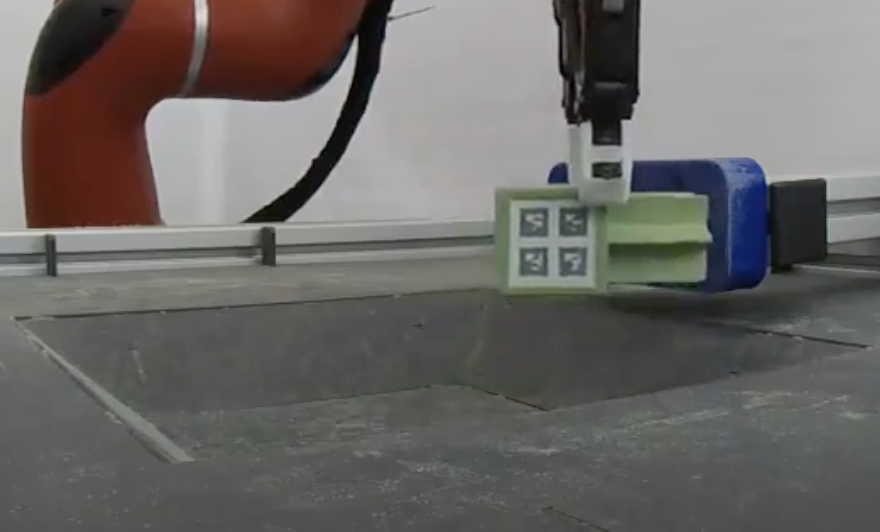}
        \caption{}
        \label{fig:s2}
    \end{subfigure}
    \begin{subfigure}{.24\textwidth}
    \centering
        \includegraphics[width=1.\linewidth]{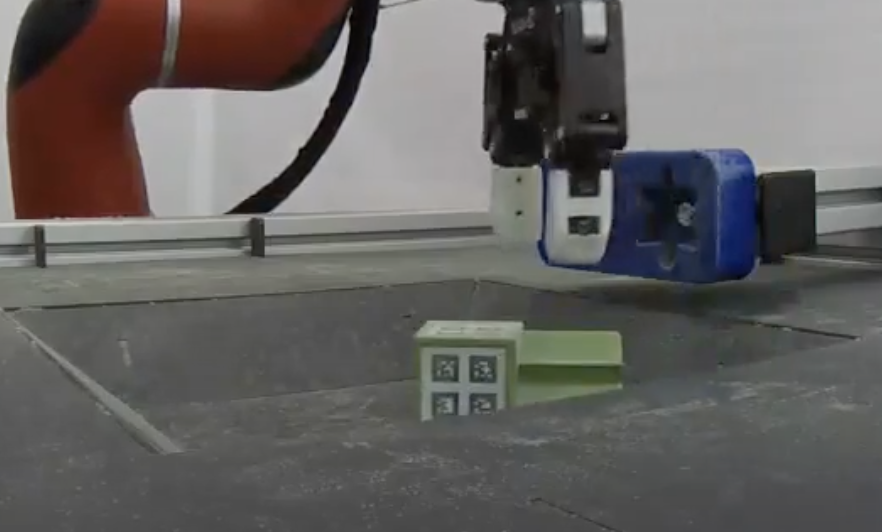}
        \caption{}
        \label{fig:s3}
    \end{subfigure}
    \begin{subfigure}{.24\textwidth}
    \centering
        \includegraphics[width=1.\linewidth]{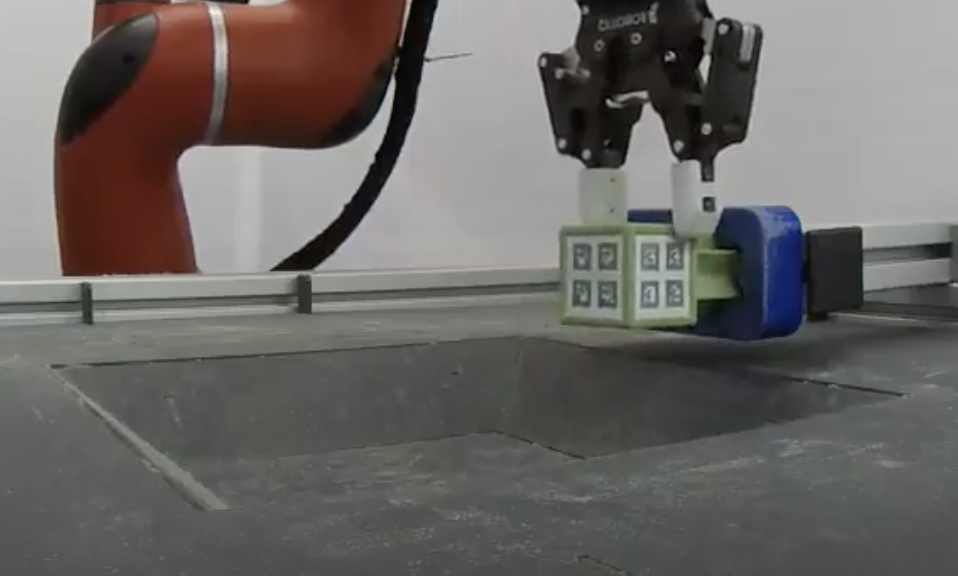}
        \caption{}
        \label{fig:s4}
    \end{subfigure}
    \caption{Reorientation by dropping. The peg is initially located in the basket with the wrong tip orientation (Fig.~\ref{fig:s1}). The robot grasps it and moves above one of the basket slopes (Fig.~\ref{fig:s2}). The robot drops the peg which ends up in the correct orientation (Fig.~\ref{fig:s3}). The robot can now grasp the peg from the top with the proper orientation for the insertion (Fig.~\ref{fig:s4}).}
    \label{fig:sequence_drop}
\end{figure}

\begin{figure}[t!]
    \centering
    \begin{subfigure}{.24\textwidth}
    \centering
          \includegraphics[width=1.\linewidth]{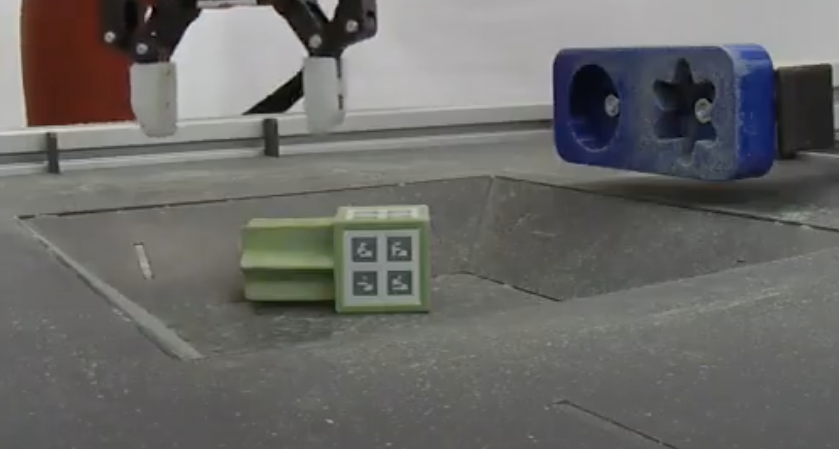}
          \caption{}
          \label{fig:s01}
    \end{subfigure}
    \begin{subfigure}{.24\textwidth}
    \centering
        \includegraphics[width=1.\linewidth]{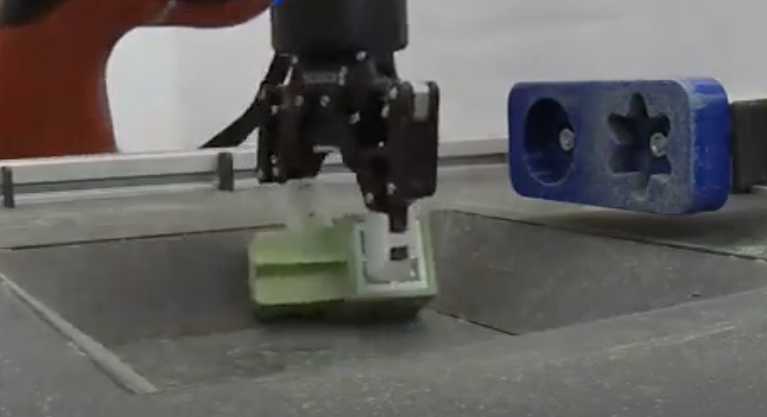}
        \caption{}
        \label{fig:s02}
    \end{subfigure}
    \begin{subfigure}{.24\textwidth}
    \centering
        \includegraphics[width=1.\linewidth]{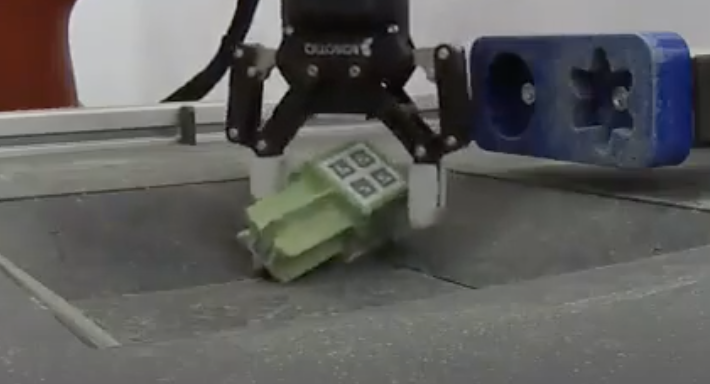}
        \caption{}
        \label{fig:s03}
    \end{subfigure}
    \begin{subfigure}{.24\textwidth}
    \centering
        \includegraphics[width=1.\linewidth]{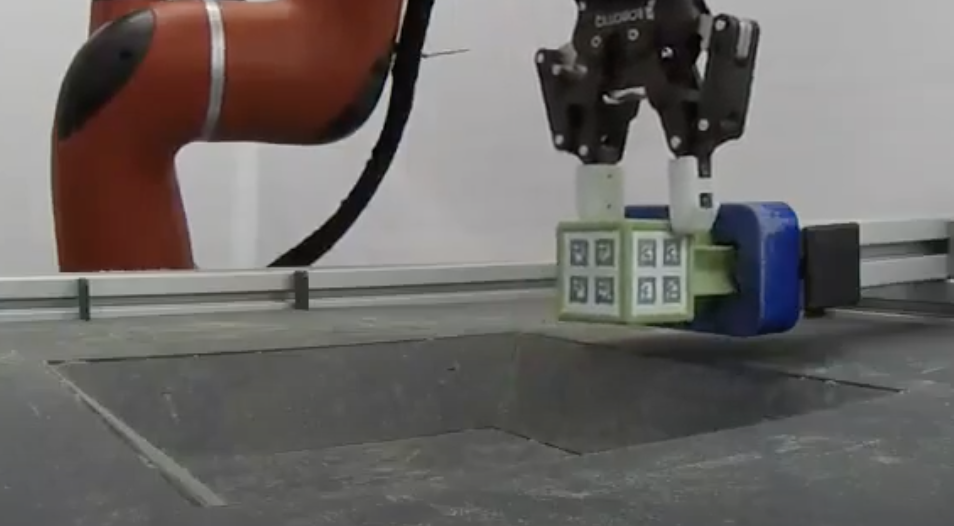}
        \caption{}
        \label{fig:s04}
    \end{subfigure}
    \caption{Reorientation by poking. The peg is initially located in the basket with the wrong tip orientation (Fig.~\ref{fig:s01}). The robot pokes one of the star corners (Fig.~\ref{fig:s02})  until the peg flips (Fig.~\ref{fig:s03}). The robot can now grasp the peg from the top with the proper tip orientation for the insertion (Fig.~\ref{fig:s04}).}
    \label{fig:sequence_corner}
\end{figure}

Our experiments show that smart behaviours concerning the tip reorientation do indeed emerge while solving the task.  In particular, the agent develops two main under-actuated strategies for that:
\begin{itemize}
    \item \textit{Reorientation by dropping}. When the peg is located on the basket with the wrong tip orientation (Fig.~\ref{fig:wrong_orient}), the agent grasps and drops the peg, if needed multiple times, until the peg flips into the correct orientation for insertion (Fig.~\ref{fig:sequence_drop}). During this behavior the peg is dropped either in the center of the basket or, preferably, on the lateral slopes, where it is more likely to flip. 
    \item \textit{Reorientation by poking}. In this case, the robot reorients the peg  by poking on the faces of the star-shaped tip (Fig.~\ref{fig:sequence_corner}). Although very impressive, this  strategy is progressively abandoned by the agent during training in favour of the dropping strategy. Poking on the peg to reorient it can yield significant forces, triggering the robot's safety and thus early terminations of the episode. Furthermore, such a strategy turns out to require in general more time to reorient the peg rather than just grasping and dropping it several time. Even if this behaviour becomes less and less executed by the agent, it is still worth noticing that multiple strategies emerged and only the most effective is utilized in the final policy.
\end{itemize}

We can therefore conclude that smart reorientation behaviours emerge from the solution of the task and that the agent can learn how to reorient the peg while facing under-actuation, even if not provided with explicit instructions to do so. Furthermore, the agent learns that correct reorientation is an essential pre-condition to insertion. We can observe that if the tip is correctly oriented, the agent quickly inserts the peg while otherwise it resorts to reorientation, if needed even multiple times. This is especially surprising given that our policy is non-recurrent.

\section{Conclusion}
\label{sec:conclusion}

In this work, we have proposed a challenging insertion task, consisting of grasping  a  free,  rotational  asymmetrical  peg,  reorienting  it  (if  needed), aligning it and final inserting it inside a rotational asymmetric hole. The task has been designed purposefully to force the agent to learn to reorient the object while facing high under-actuation.
As a result of only providing reward for success (\emph{what}) but not specifying \emph{how} to solve the task, the agent develops interesting under-actuated reorientation strategies that were not considered during the design of the task, such as reorienting the peg by grasping and dropping it or by poking the tip in order to flip the peg. Furthermore, the agent correctly learns to identify these behaviors as a pre-condition to solving the task, in-spite of the long time horizon of several hundred steps.
Despite the complex policy to be learned, the task is solved from scratch, with impressive performance, both in simulation and on a single real robot. The high data efficiency results from using a multi-task RL framework, based on Scheduled Auxiliary Control (SAC-X)~\cite{riedmiller2018learning} combined with Regularized Hierarchical Policy Optimization (RHPO)~\cite{wulfmeier2019compositional}.

The results presented in this paper suggest several perspectives for future work. For instance, the multi-task formulation used in this paper could be applied to other complex manipulation tasks to learn further emerging behaviors. It could be useful to store the skills, even those who disappeared over the course of the training -- as happened for \enquote{reorientation by poking} --, and transfer them to other tasks in order to speed up learning.
Furthermore, given the great results we achieved when using reloaded (and therefore batch) data, a natural future direction for this work is to investigate if Batch RL algorithms such as~\cite{siegel2020keep} can speed up the training even further.



\acknowledgments{The authors would like to thank the DeepMind robotics team for support for the robotics experiments, with a special mention to Federico Casarini for his help in designing and implementing the real robot setup.}
\clearpage

\bibliography{main}  

\newpage
\appendix
\section{Real robot setup}
\label{appendix:robot}
The  real world setup for the insertion task consists of a Sawyer robot arm developed by Rethink Robotics.
It is equipped with a Robotiq 2F85 gripper as well as a Robotiq FTS300 wrist force torque sensor at the
wrist. In front of the robot, there is a basket with a base size of 20x20 cm and lateral slopes. The
robot is controlled in Cartesian velocity space with a maximum speed of 0.05 m/s. The arm control mode
has four control inputs: three Cartesian linear velocities as well as a rotational velocity of the wrist
around the vertical axis. Together with the gripper opening angle, the total size of the action space is
five (Table~\ref{tab:actions}).

\begin{figure}
    \centering
    \begin{subfigure}{.4\textwidth}
    \centering
          \includegraphics[width=.8\linewidth]{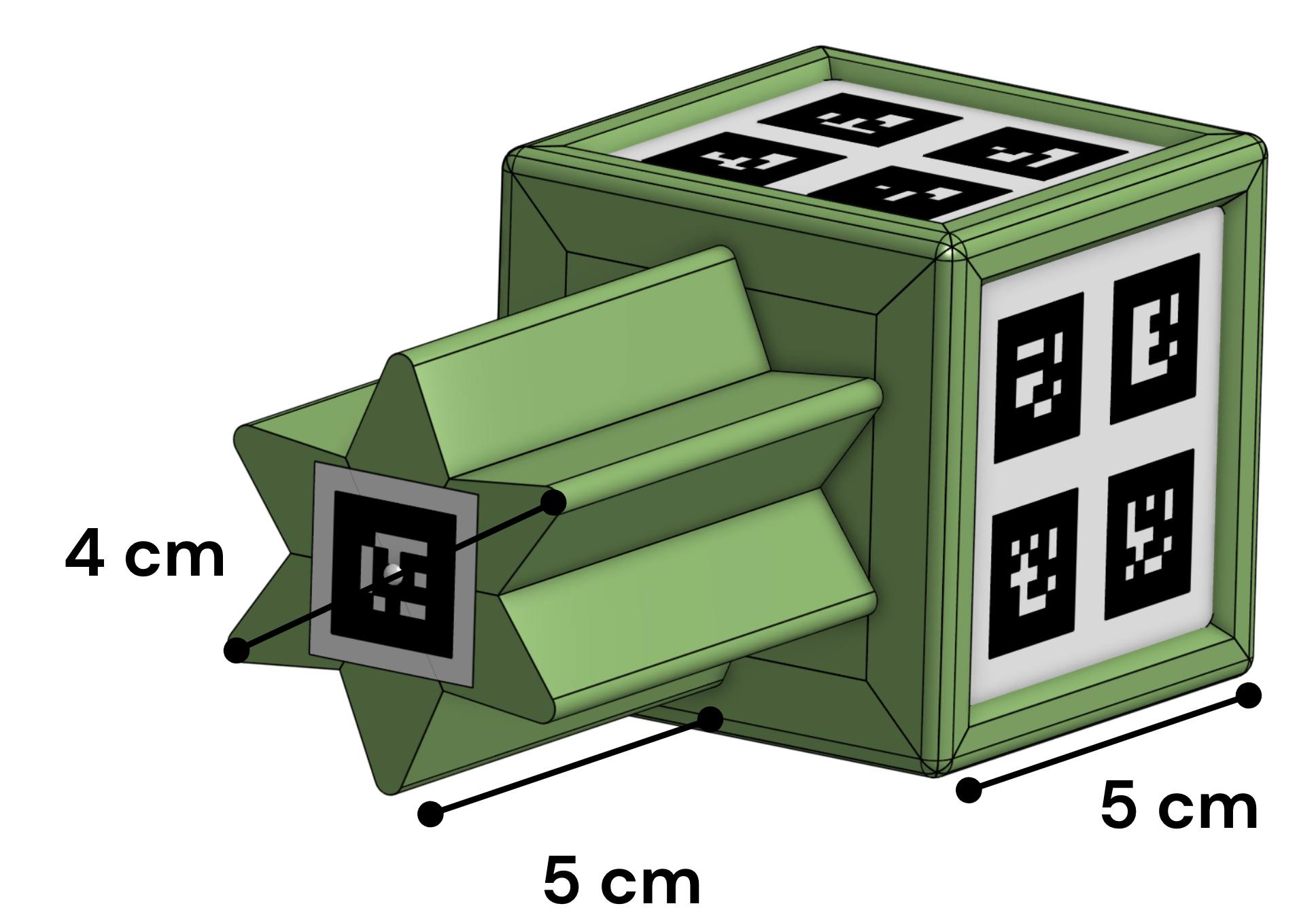}
          \caption{}
          \label{fig:peg_dimensions}
    \end{subfigure}
    \begin{subfigure}{.4\textwidth}
    \centering
        \includegraphics[width=0.9\linewidth]{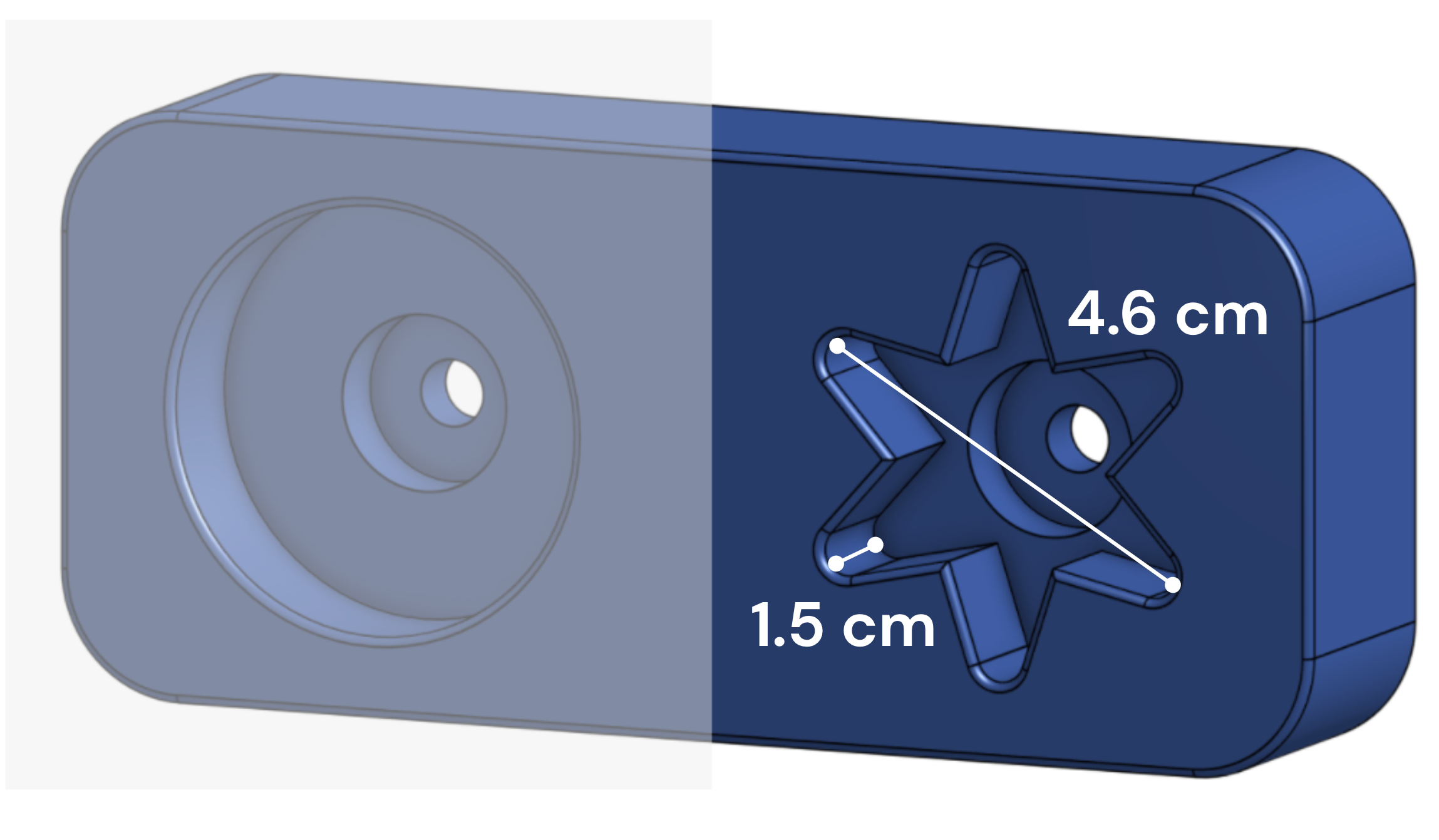}
        \caption{}
        \label{fig:hole_dimensions}
    \end{subfigure}
    \caption{Dimensions of the peg and hole used in the insertion task.}
    \label{fig:dimensions}
\end{figure}

The peg and hole dimensions are shown in Fig.~\ref{fig:dimensions}. The peg is equipped with Augmented
Reality markers on its surface. The markers are tracked by three cameras (two in front, one in the
back of the basket). Using an extrinsic calibration of the cameras, the individual measurements are fused
into a single pose estimate for the peg which is provided as an observation to the agent. 
The agent is run at 20 Hz and, together with the peg pose, receives proprioception observations from the arm (joint and end-effector
positions and velocities), the gripper (joint positions, velocities and grasp signal), wrist force-torque
sensor (force and torques) and the last executed action (Table~\ref{tab:obs}). In order to deal with any possible missing observations at a single timestep (especially the lack of the peg's linear and angular velocity), the agent receives a stack of the last three observations and actions. A successful insertion is computed based on the known fixed position of the hole and the pose estimate of the peg, so no further observations are required for that.
During all experiments, episodes are 600 steps long but are terminated early when the force-torque
measurements at the wrist exceed 15 N on any of the principal axes, to prevent damage to the setup.

\begin{table}[]
    \centering
    \begin{tabular}{c|c | c}
        Entry & Dimension & Unit \\
        \hline
        End-effector Cartesian linear velocity   & 3  & m/s \\
        Gripper rotational velocity   & 1 & rad/s \\
        Gripper finger velocity & 1 & rad/s\\
        \hline\\
    \end{tabular}
    \caption{Robot action space used both in simulation and on the real robot.}
    \label{tab:actions}
\end{table}

\begin{table}[]
    \centering
    \begin{tabular}{c|c | c}
        Entry & Dimension & Unit \\
        \hline
        Joint Angle   & 7  &rad \\
        Joint Velocity  & 7 & rad/s \\
        Joint Torque & 7 & N$\cdot$m\\
        Wrist Force & 3 & N \\
        Wrist Torque & 3 & N$\cdot$m\\
        Pinch Pose &7& m, au\\
        Finger angle & 1 & rad\\
        Finger velocity & 1 & rad/s\\
        Grasp & 1 & n/a\\
        Peg Pose &7& m, au\\
        Peg pose wrt Pinch & 7 & m, au \\
        Action & 5 & m/s, rad/s\\
        \hline\\
    \end{tabular}
    \caption{Observations used in the real robot experiments. In the table au stands for  arbitrary units used for the quaternion representation.}
    \label{tab:obs}
\end{table}

\begin{table}[]
    \centering
    \begin{tabular}{c|c | c}
        Entry & Dimension & Unit \\
        \hline
        Joint Angle   & 7  &rad \\
        Joint Velocity  & 7 & rad/s \\
        Pinch Pose &7& m, au\\
        Finger angle & 1 & rad\\
        Finger velocity & 1 & rad/s\\
        Grasp & 1 & n/a\\
        Peg Pose &7& m, au\\
        Peg pose wrt Pinch & 7 & m, au \\
        \hline\\
    \end{tabular}
    \caption{Observations used in simulation. In the table au stands for  arbitrary units used for the quaternion representation.}
    \label{tab:obs_sim}
\end{table}

\section{Simulated environment}
\label{appendix:sim}
For simulating the real robot setup we use the physics
simulator MuJoCo~\citep{mujoco}.  The simulation is run with a physics integration time step of 0.5
milliseconds, with control interval of 50 milliseconds (20 Hz) for the agent. The simulated environment has been designed to replicate the kinematics of the real robot setup by using the peg and hole meshes from the CAD in simulation that are also used for the 3D printing the real peg and hole objects. However it should be noted that accurate simulation of forces and contact dynamics are very challenging to achieve and hence the forces and torques applied to the wrist of the robot are not modelled accurately and nor provided to the agent as part of the observations.

\section{Shaped reward design}
\label{appendix:shaped_reward}
\begin{figure}
    \centering
    \begin{subfigure}{.32\textwidth}
    \centering
          \includegraphics[width=.8\linewidth]{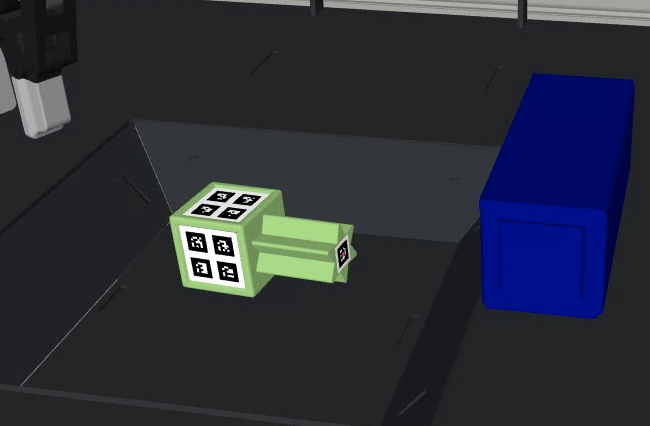}
          \caption{}
          \label{fig:correct1}
    \end{subfigure}
    \begin{subfigure}{.33\textwidth}
    \centering
          \includegraphics[width=.81\linewidth]{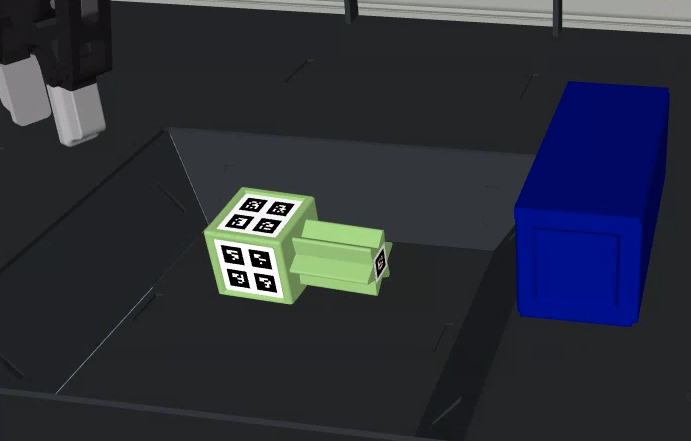}
          \caption{}
          \label{fig:correct2}
    \end{subfigure}
    \begin{subfigure}{.32\textwidth}
    \centering
        \includegraphics[width=0.88\linewidth]{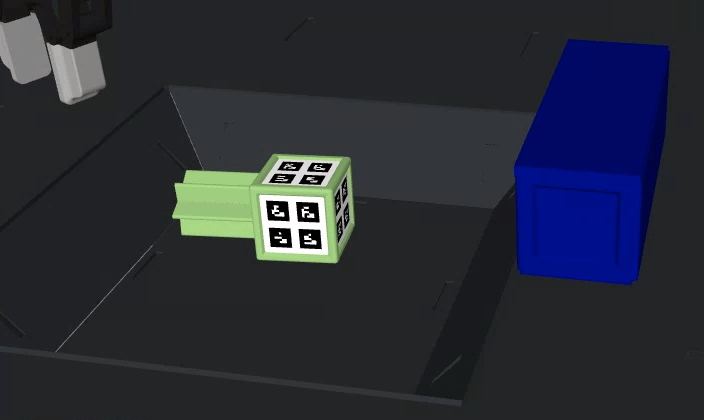}
        \caption{}
        \label{fig:wrong}
    \end{subfigure}
    \caption{Examples of different peg  alignments. In Fig.~\ref{fig:correct1}, the peg is properly aligned to the direction of insertion and  $R_{align}=1$. The same happens in Fig.~\ref{fig:correct2}: even if the peg's tip is in the wrong orientation (see the star orientation), the peg is properly aligned to the direction of insertion and, therefore, $R_{align}=1$. In Fig.~\ref{fig:wrong} instead the peg is not aligned to the direction of insertion, and in this case $R_{align}=0$.}
    \label{fig:alignment}
\end{figure}

The shaped reward we use in the MPO experiments (Section~\ref{sec:mpo}) is designed by taking inspiration from the rewards proposed in~\cite{riedmiller2018learning}.  The reward sums up several skill rewards:
\begin{equation}
\label{eq:shaped_reward_appendix}
    R = \frac{1}{7} [R_{reach} + R_{open} + R_{grasp} + R_{lift} + R_{push} + R_{align} + R_{insert}],
\end{equation}
where each individual term is normalized to a maximum of 1 when:
\begin{itemize}
    \item   The gripper reaches the peg ($R_{reach}$):
       \begin{equation}
           \label{eq:reach_reward}
        R_{reach} = 
        \begin{cases}
        1 & \text{iff } \lVert\frac{p_{gripper} - {p_{peg}}}{tol_{pos}}\rVert \leq 1,\\
        1 - tanh^2(\lVert\frac{p_{gripper} - {p_{peg}}}{tol_{pos}}\rVert \cdot \epsilon),&  \text{otherwise}.\\
        \end{cases}
       \end{equation}
    $p_{gripper} \in \mathbb{R}^3$ is the position of the gripper, $p_{peg} \in \mathbb{R}^3$ the position of the handle of the peg, $tol_{pos} = [0.055, 0.055, 0.02]$ m the tolerance in position and $\epsilon$ a scaling factor.
    \item  The gripper fingers are open ($R_{open}$)\footnote{We also tried removing $R_{open}$ from the shaped reward of Eq.~\ref{eq:shaped_reward_appendix} because it might be in contrast with the grasp reward of Eq.~\ref{eq:grasp_reward} ($R_{grasp}$). However, this does not improve the performance and even in that case MPO is not able to properly solve the task. }:
    \begin{equation}
    \label{eq:open_reward}
        R_{open} = 
        \begin{cases}
        1 & \text{iff } \lVert\frac{p_{finger} - {p_{desired}}}{tol_{pos}}\rVert \leq 1,\\
        1 - tanh^2(\lVert\frac{p_{finger} - {p_{desire}}}{tol_{pos}}\rVert \cdot \epsilon),&  \text{otherwise}.\\
        \end{cases}
    \end{equation}
    $p_{finger} \in \mathbb{R}$ is the angle of the gripper fingers, $p_{desired} \in \mathbb{R}$ is the angle corresponding to open fingers, $tol_{pos} = 1e^{-9}$ is the tolerance in position and $\epsilon$ a scaling factor.
    \item  The robot is grasping the peg ($R_{grasp}$):
    \begin{equation}
    \label{eq:grasp_reward}
        R_{grasp} =  0.5 ( R_{closed} + R_{grasp, aux}), 
    \end{equation}
    where:
    \begin{equation}
    \label{eq:close_reward}
        R_{closed} = 
        \begin{cases}
        1 & \text{iff } \lVert\frac{p_{finger} - {p_{desired}}}{tol_{pos}}\rVert \leq 1,\\
        1 - tanh^2(\lVert\frac{p_{finger} - {p_{desire}}}{tol_{pos}}\rVert \cdot \epsilon),&  \text{otherwise}.\\
        \end{cases}
    \end{equation}
    with the quantities having the same meaning as in Eq.~\ref{eq:open_reward}, except for $p_{desired} \in \mathbb{R}$, which is the angle corresponding to closed fingers  this time;\\
    \begin{equation}
    \label{eq:grasp_sensor_reward}
        R_{grasp,aux} = 
        \begin{cases}
        1 & \text{If grasp is detected by the gripper sensor,} \\
        0&  \text{otherwise}.\\
        \end{cases}
    \end{equation}
    \item  The peg is lifted above a certain height ($R_{lift}$):
    \begin{equation}
    \label{eq:lift_reward}
        R_{lift} = 
        \begin{cases}
        1 & \text{iff } z_{peg} \geq z_{desired},\\
        \frac{z_{peg} - z_{min}}{z_{desired} - z_{min}}&  \text{otherwise}.\\
        \end{cases}
    \end{equation}
    $z_{peg}$ is the z-coordinate of the peg, $z_{desired} = 0.18$ m is the desired peg z-coordinate, $z_{min} = 0.08$ m is the minimum peg z-coordinate. 
    \item  The peg has been pushed close to the hole ($R_{push}$):
    \begin{equation}
    \label{eq:push_reward}
        R_{push} = 
        \begin{cases}
        1 & \text{iff } \lVert\frac{p_{tip} - {p_{hole}}}{tol_{insert}}\rVert \leq 1,\\
        1 - tanh^2(\lVert\frac{p_{tip} - {p_{hole}}}{tol_{insert}}\rVert \cdot \epsilon),&  \text{otherwise}.\\
        \end{cases}
    \end{equation}
    $p_{tip} \in \mathbb{R}^3$ is the position of the tip of the peg (computed by using the position of the centre of the peg, the peg orientation and dimensions), $p_{hole} \in \mathbb{R}^3$ is the position of the centre of the hole, $tol_{insert} = [0.02, 0.02, 0.02]$ m is the tolerance in position and $\epsilon$ a scaling factor.
    \item  The peg is aligned to the axis of insertion ($R_{align}$):
    \begin{equation}
    \label{eq:align_reward}
        R_{align} = 1 - \frac{\alpha}{\pi},
    \end{equation}
    where $\alpha$ is the angle between the axis along the longest dimension of the peg (roll axis) and the desired direction of insertion.
    Note that this does not provide any information on the proper orientation of the peg's tip. This reward just encourages the robot to align the peg along the direction of insertion (see Fig.~\ref{fig:alignment}), i.e. to rotate the peg only around its yaw axis, and does not provide any information on the required rotation around its roll axis.
    \item  The peg is inside the hole (and the robot is holding it) ($R_{insert}$):
    \begin{equation}
        R_{insert} = R_{insert, dist} \cdot R_{insert, align}, 
        \label{eq:insert}
    \end{equation}
    with:
    \begin{equation}
    \label{eq:insert_dist}
        R_{insert, dist} = 
        \begin{cases}
        1 & \text{iff } \lVert\frac{p_{tip} - {p_{hole}}}{tol_{insert}}\rVert \leq 1,\\
        0 &  \text{otherwise}.\\
        \end{cases}
    \end{equation}
    where $p_{tip} \in \mathbb{R}^3$ is the position of the tip of the peg (computed by using the position of the centre of the peg, the peg orientation and dimensions), $p_{hole} \in \mathbb{R}^3$ is the position of the centre of the hole, $tol_{insert} = [0.005, 0.005, 0.005]$ m is the tolerance required for the insertion.
    \begin{equation}
    \label{eq:insert_align}
        R_{insert, align} = 
        \begin{cases}
        1 & \text{iff } (1 - \frac{\alpha}{\pi}) \geq 0.8,\\
        0 &  \text{otherwise}.\\
        \end{cases}
    \end{equation}
    with $\alpha$ the angle between the axis along the longest dimension of the peg and the desired direction of insertion.
    Even if not explicitly required by the insertion reward, the depth of the hole and the length of the peg are such that the peg can stay inside the hole only if the robot keeps holding it.
    
\end{itemize}
This reward strongly encourages a sequence of steps to be executed by the robot. For example, if the robot manages to reorient the peg but then its orientation changes while grasping it, this is penalized by the reward.

\section{Multi-task formulation}
\label{appendix:multi_task}
\begin{figure}
    \centering
    \begin{subfigure}{.32\textwidth}
    \centering
          \includegraphics[width=.8\linewidth]{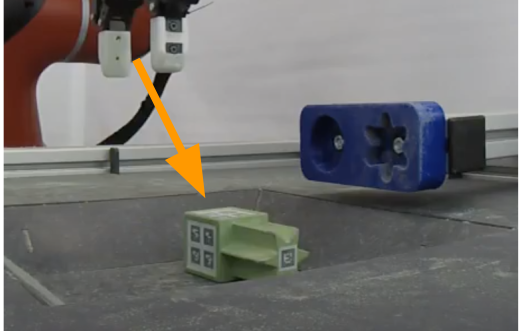}
          \caption{Reach intention.}
          \label{fig:reach_int}
    \end{subfigure}
    \begin{subfigure}{.33\textwidth}
    \centering
          \includegraphics[width=.79\linewidth]{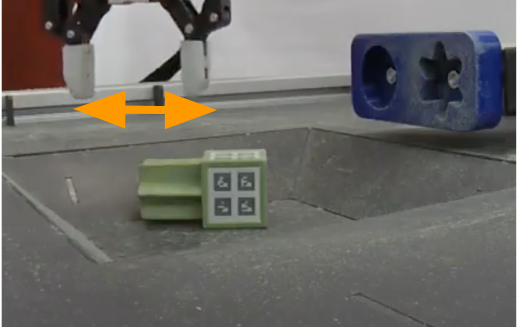}
          \caption{Open intention.}
          \label{fig:open_int}
    \end{subfigure}
    \begin{subfigure}{.32\textwidth}
    \centering
        \includegraphics[width=.84\linewidth]{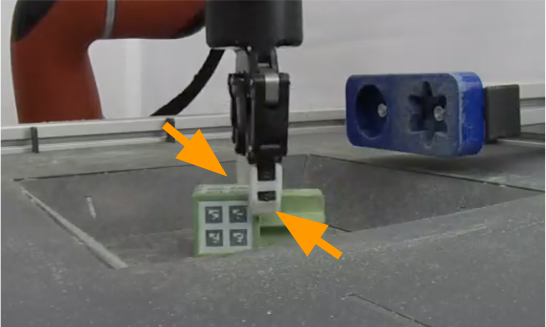}
        \caption{Grasp intention.}
        \label{fig:grasp_int}
    \end{subfigure}
    \begin{subfigure}{.32\textwidth}
    \centering
        \includegraphics[width=.8\linewidth]{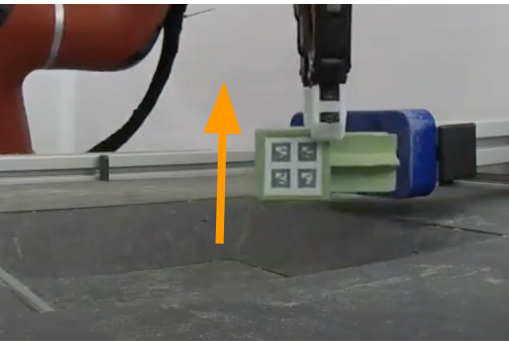}
        \caption{Lift intention.}
        \label{fig:lift_int}
    \end{subfigure}
    \begin{subfigure}{.32\textwidth}
    \centering
        \includegraphics[width=.83\linewidth]{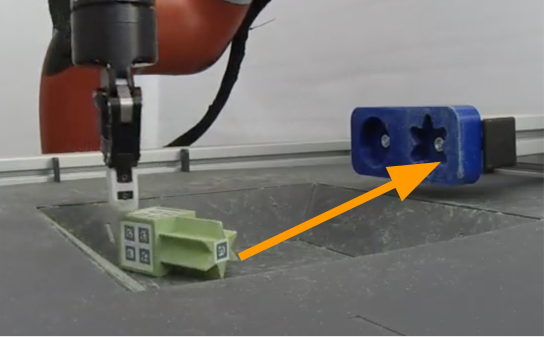}
        \caption{Push intention.}
        \label{fig:push_int}
    \end{subfigure}
    \begin{subfigure}{.32\textwidth}
    \centering
        \includegraphics[width=.8\linewidth]{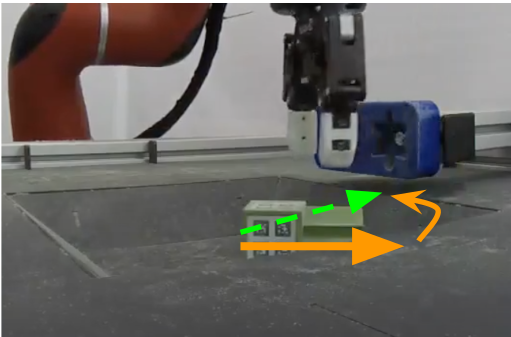}
        \caption{Align intention.}
        \label{fig:align_int}
    \end{subfigure}
    \begin{subfigure}{.5\textwidth}
    \centering
        \includegraphics[width=.55\linewidth]{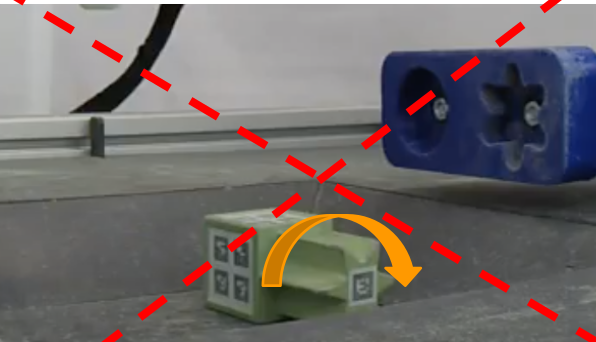}
        \caption{No reorientation intention is used.}
        \label{fig:no_reorientation}
    \end{subfigure}
    \caption{ Graphical explanation of the auxiliary intentions used within the SAC-X framework (Figs.~\ref{fig:reach_int} - \ref{fig:align_int}). No reward instead is provided for the reorientation of the peg's tip around the roll axis (Fig.~\ref{fig:no_reorientation}).}
    \label{fig:intentions_images}
\end{figure}

In order to use SAC-X on our insertion task, we provide the agent a set of auxiliary intentions, that it attempts to learn simultaneously via off-policy RL together with the main insertion task. The main goal of such a formulation is to help the agent during exploration of the environment.
The auxiliary intentions we design correspond to the skills potentially useful for solving the main task, i.e. reach, grasp, lift, push, align the peg and open the fingers. The rewards we use for the skills are the ones from Eqs. \ref{eq:reach_reward} - \ref{eq:align_reward} presented in Section~\ref{appendix:shaped_reward}. For the sake of clarity, Fig.~\ref{fig:intentions_images} provides a graphical explanation of the auxiliary intentions. The sparse insertion reward, defined in Eq.~\ref{eq:insert}, is included in the SAC-X framework as the main objective. Remarkably, we do not include any intention to encourage the robot to reorient the peg's tip (see Fig.~\ref{fig:no_reorientation}).

SAC-X provides the options to use a random scheduler (SAC-U) or a learned scheduler (SAC-Q)~\cite{riedmiller2018learning}. During our experiments we also train the scheduler, using therefore SAC-Q. The episode length is 600 steps and the scheduler can execute up to three different intentions within an episode.

\section{Training details and hyperparameters}

All the trainings reported within this work has been executed from features, being the state of the system available (in simulation and, partially, on the real robot) or estimated from sensor measurements (on the real robot).
The simulation experiments have been executed on 3 different random seeds. Given the time required to train the task on the real robot, only one experiment of each type has been carried out on the real robot.
The evaluation on the real robot has executed by applying only the mean of the policy.

\subsection{MPO}
\label{appendix:mpo_params}
\subsubsection{Networks architectures}
The networks used within the MPO experiments are feedforward networks with first layer normalization and ELU activations. More details in Table~\ref{tab:mpo_params}. 
\subsubsection{Algorithm Hyperparameters}
The hyperparameters used in the MPO experiments both in simulation and on the real robot are reported in Table~\ref{tab:mpo_params}.
\begin{table}[ht!]
    \centering
    \begin{tabular}{c|c}
        Hyperparameter & Value \\
        \hline
        Policy net & 256 - 256 \\
        Q function net &  256 - 256\\
        $\epsilon$ & 0.1 \\
        $\epsilon_{\mu}$ & 0.005 \\
        $\epsilon_{\Sigma}$ & 0.00001 \\
        Number of action samples & 20\\
        Discount factor & 0.99\\
        Replay buffer size & 2$e^6$ \\
        Target network update period & 200 \\
        Batch size & 512 \\
        Adam learning rate & $3e^{-4}$\\
        \hline\\
    \end{tabular}
    \caption{Hyperparameters used both for the simulated  and real  MPO experiments.}
    \label{tab:mpo_params}
\end{table}

\subsection{SAC-X combined with RHPO}
\subsection{Networks architectures}
\begin{figure}
    \centering
    \begin{subfigure}{.4\textwidth}
        \centering
      \includegraphics[width=.7\linewidth]{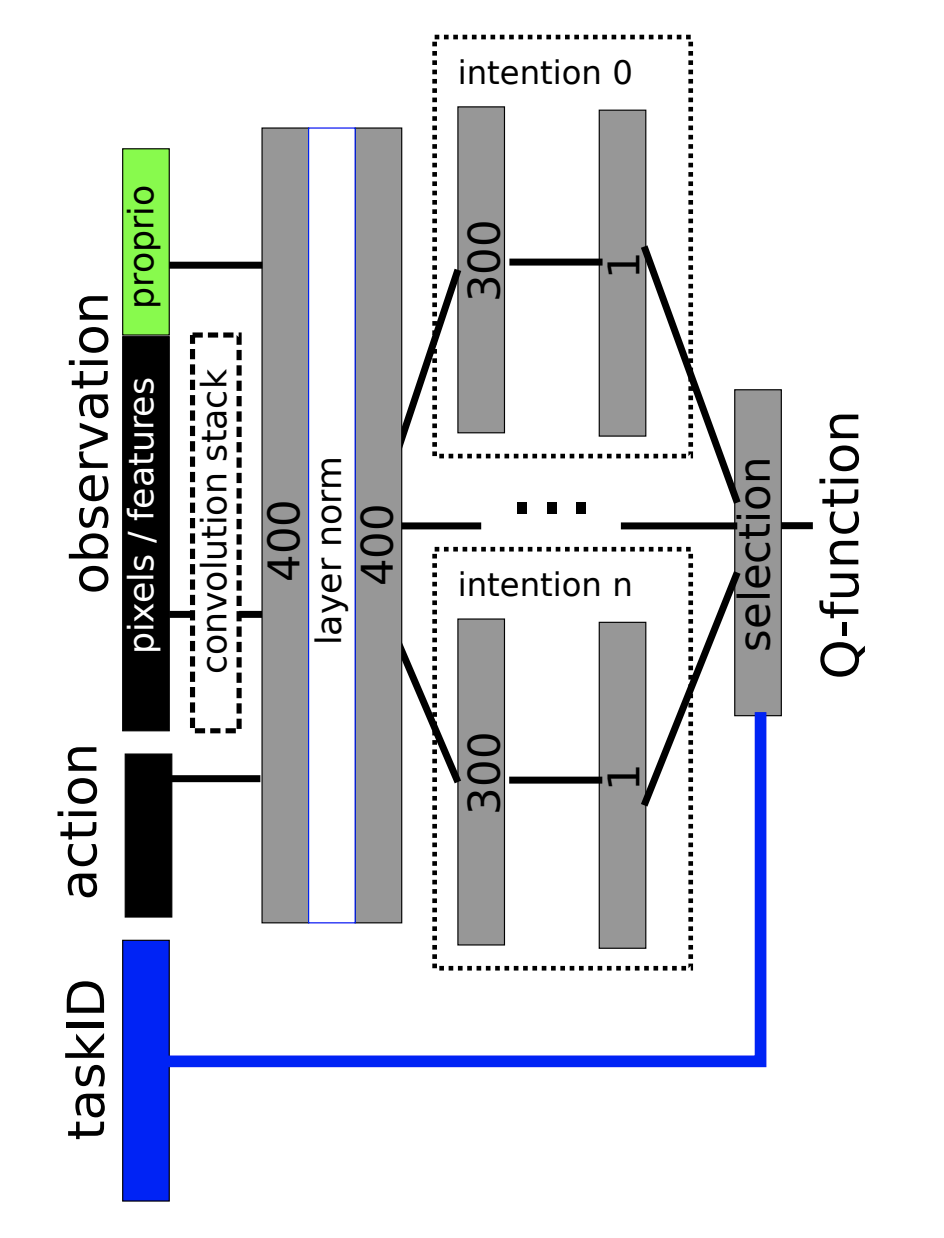}
      \caption{}
      \label{fig:q_net}
    \end{subfigure}
     \begin{subfigure}{.4\textwidth}
     \centering
      \includegraphics[width=1.\linewidth]{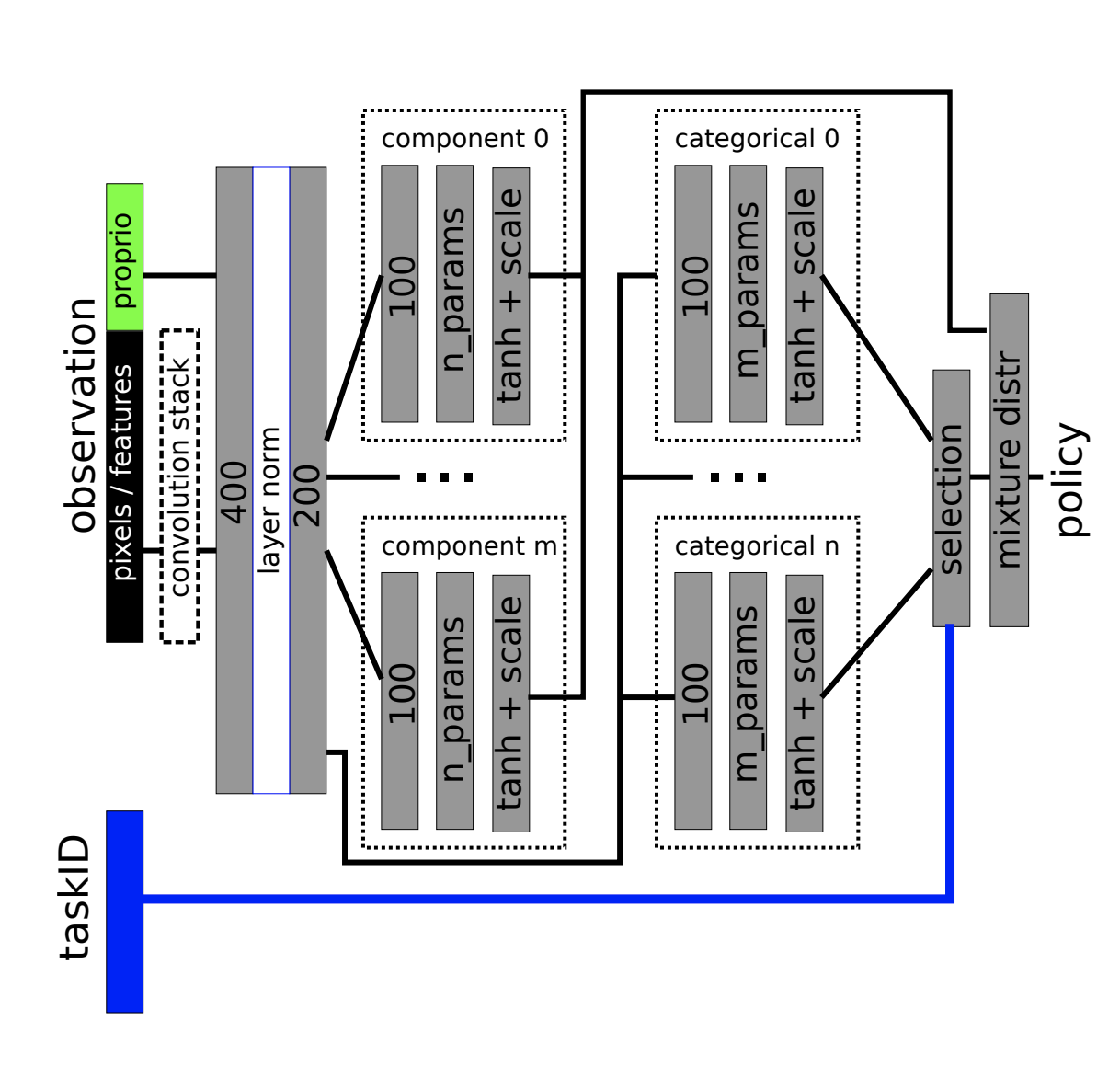}
      \caption{}
      \label{fig:pi_net}
    \end{subfigure}
    \caption{ Q-function and policy networks used in SAC-X + RHPO experiments both in simulation and on the real robot.}
    \label{fig:nets}
\end{figure}

To represent the Q-function we use the network architecture proposed in~\cite{riedmiller2018learning} (Fig.~\ref{fig:q_net}). The proprioception of the
robot, the features of the objects and the actions are fed together into a torso network. At the input we use a fully connected first layer of 400 units, followed by a layer normalization operator, a tanh activation and another fully connected layer of 400 units with an ELU activation function. The output of
this torso network is shared by independent head networks
for each of the intentions. Each head has two fully connected layers and outputs a Q-value for this task, given the input of the network. Using the task identifier we then can compute the Q value
for a given sample by discrete selection of the according head output.

The policy architecture we use is instead the one proposed in~\cite{wulfmeier2019compositional} (Fig.~\ref{fig:pi_net}). The network consists of a torso and
a set of networks parameterizing the Gaussians distributions (named components in (Fig.~\ref{fig:pi_net})) which are 
shared across tasks and a task-specific network to parameterize the categorical distribution for each task (named categoricals in (Fig.~\ref{fig:pi_net})). The final policy
distribution is task-dependent for the high-level controller but
task-independent for the low-level policies.

\subsubsection{Algorithm hyperparameters}
The hyperparameters used in the SAC-X + RHPO experiments both in simulation and on the real robot are reported in Table~\ref{tab:rhpo_params}.

\begin{table}[ht!]
    \centering
    \begin{tabular}{c|c}
        Hyperparameter & Value \\
        \hline
        Policy torso (shared across tasks)    & 400 - 200\\
        Policy task-dependent head & 100 (controllers) \\
        Policy shared head & 100 (components) \\
        Q function torso  (shared across tasks) & 400 - 400 \\
        Q function head (per task) & 300\\ 
        Number of action sample & 20 \\
        Number of components &  3 \\
        Discount factor & 0.99\\
        Replay buffer size & 1$e^6\times$ number of tasks \\
        Target network update period & 500 \\
        Batch size & 512 \\
        \hline\\
    \end{tabular}
    \caption{Hyperparameters used both for the simulated  and real experiments with SAC-X + RHPO.}
    \label{tab:rhpo_params}
\end{table}

\subsubsection{Simulation vs real robot experiments}
\label{sec:sim_vs_real}
The plots shown in Fig.~\ref{fig:sim_real_plots} in the main paper require some explanation to fairly compare the results obtained in simulation to those obtained with the real robot. From a first comparison, it might seem that, in simulation, 10 times more episodes are required to achieve the same performance we obtained on the real robot. Hereafter, we explain the reasons behind this gap:
\begin{itemize}
    \item The rate at which data is generated in simulation and on the real robot is different. In particular, using multiple actors (64 in our case)\footnote{We used multiple actors instead of a single actor because rolling an episode in simulation is slower than on the real robot.} in a distributed system leads to a large quantity of data generated. Therefore, the learner saturates and uses the same data a few  times. In the single real robot case instead, the actor is the bottleneck. Data is generated slower and therefore the learner reuses the same data more times. In our case in particular we have that the ratio between the data used by the learner and that generated by the actor in simulation is half the same ratio on the real robot. For this reason, we can conclude that in simulation the task is solved actually \textit{5 times} (and not 10) slower.
    \item The simulation task turns out to be harder than the one on the real robot.  The task involves a lot of physical interaction with objects and between the objects and the environment. The difficulty in simulating these behaviours sometimes leads to unrealistic behaviours of the object which make the task harder.
\end{itemize}

\section{Further experiment plots}

\subsection{MPO with shaped reward}
\label{appendix:mpo}
\begin{figure}
    \centering
    \begin{subfigure}{.45\textwidth}
    \centering
          \includegraphics[width=1.\linewidth]{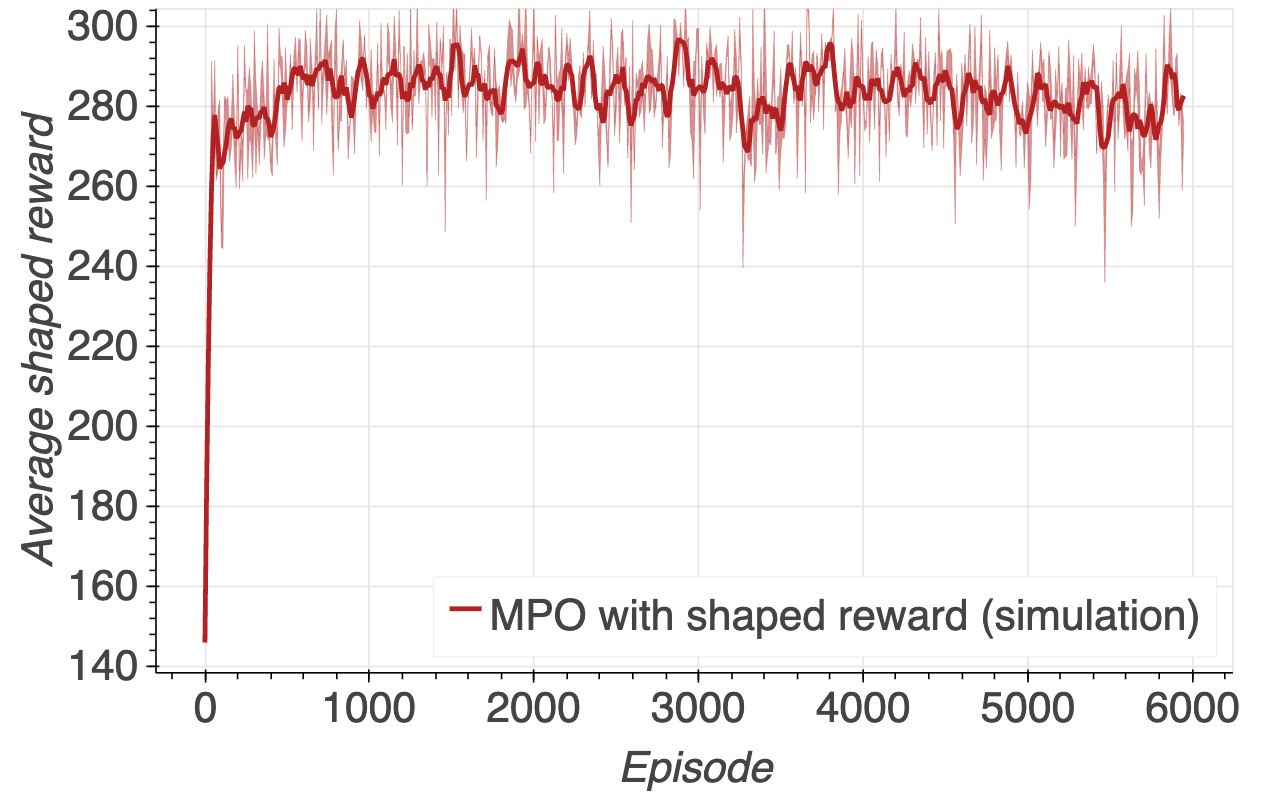}
          \caption{}
          \label{fig:sim_plot_mpo}
    \end{subfigure}
    \begin{subfigure}{.45\textwidth}
    \centering
        \includegraphics[width=1.\linewidth]{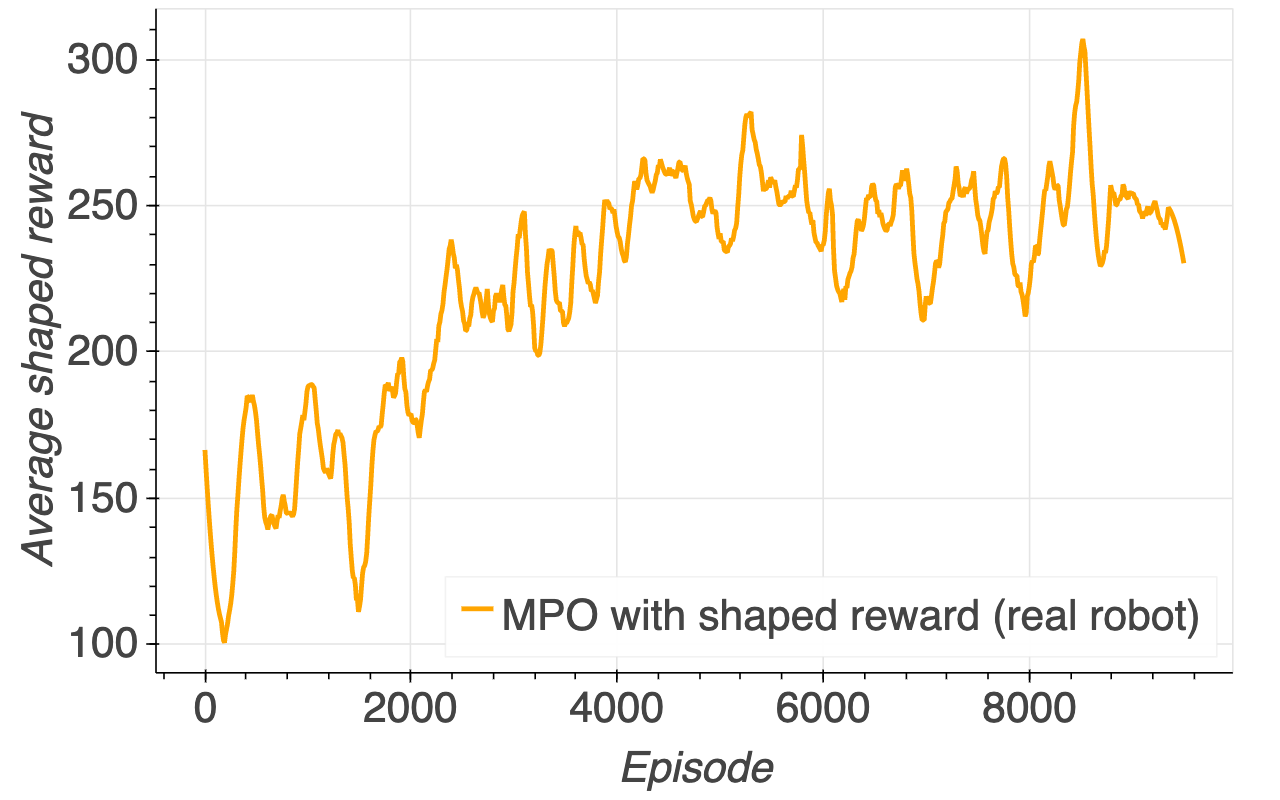}
        \caption{}
        \label{fig:real_plot_mpo}
    \end{subfigure}
    \caption{MPO averaged shaped reward  for simulation (Fig.~\ref{fig:sim_plot}) (window size: 10) and real robot experiments (window size: 50). The real robot experiments have been run for 10 days, without leading to any successful insertions. (Fig.~\ref{fig:real_plot})}.
    \label{fig:sim_real_plots_mpo}
\end{figure}

MPO with the shaped reward is not able to learn the task. Figs.~\ref{fig:sim_plot_mpo} and \ref{fig:real_plot_mpo} show the reward experienced during the training respectively in simulation and on the real robot.  Unlike the results obtained with SACX combined with RHPO (see Figs.\ref{fig:real_plot}, \ref{fig:reloaded_plot}), a non-zero reward in this experiment does not represent the solution of the task, since Eq.~\ref{eq:shaped_reward_appendix} always provide non-zero values. The minimum value for which the task can be considered solved using this reward is 400. The task episode length is equal to 600 and, in a successful episode, five over seven skills (reach, push, lift, grasp, orient) have to be solved since almost the beginning of the episode and therefore provide reward equal to 1. This leads to a minimum reward equal tp $5/7 * 600 \approx 400$ for successful episodes.
\subsection{SAC-X combined with RHPO - from scratch}
\begin{figure}
    \centering
    \begin{subfigure}{.3\textwidth}
        \centering
      \includegraphics[width=1.\linewidth]{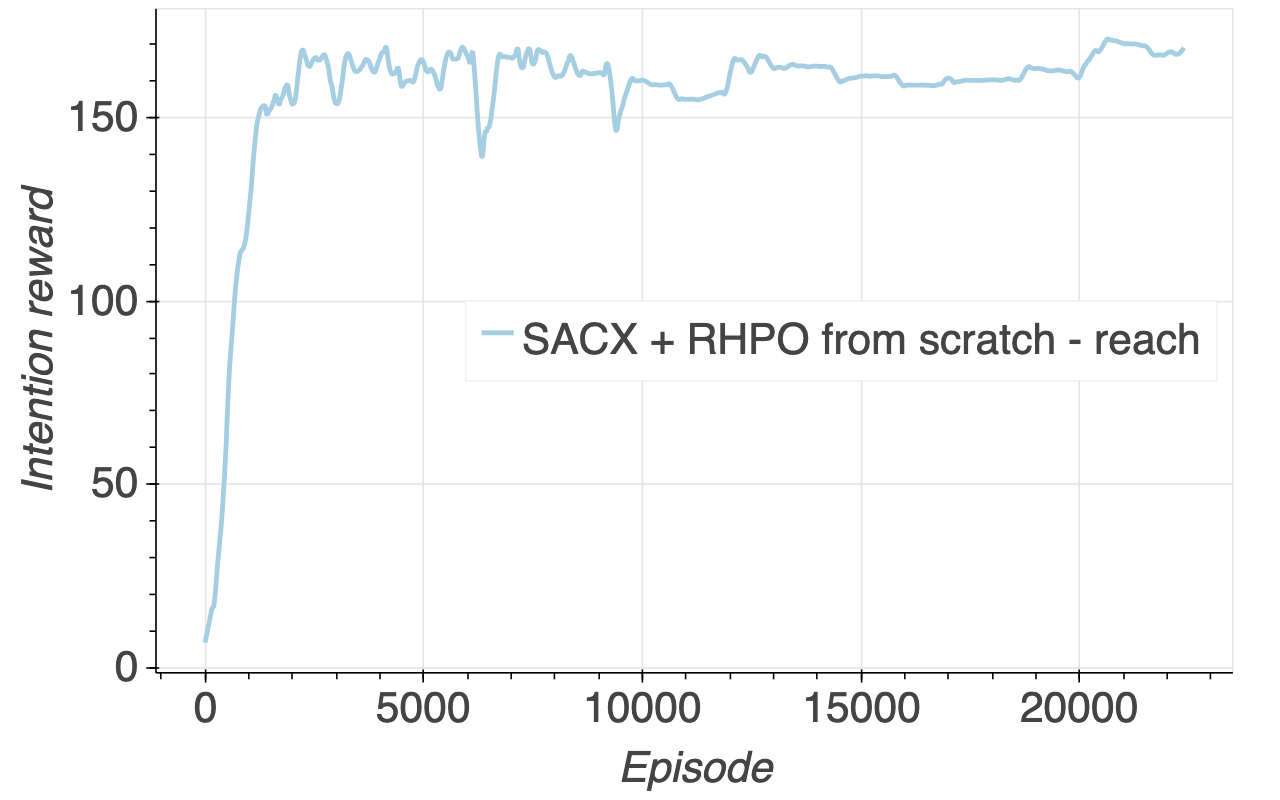}
      \caption{}
      \label{fig:1_scratch}
    \end{subfigure}
     \begin{subfigure}{.3\textwidth}
     \centering
      \includegraphics[width=1.\linewidth]{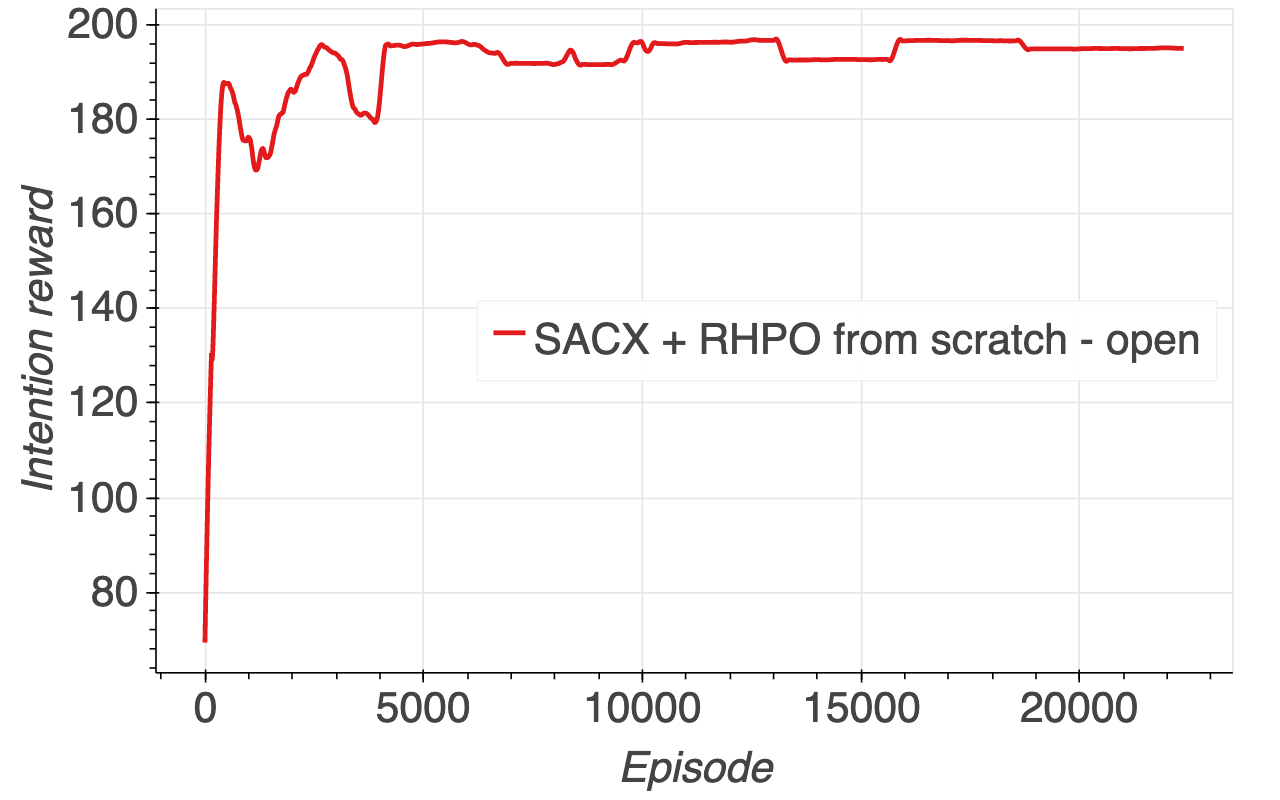}
      \caption{}
      \label{fig:2_scratch}
    \end{subfigure}
    \begin{subfigure}{.3\textwidth}
     \centering
      \includegraphics[width=1.\linewidth]{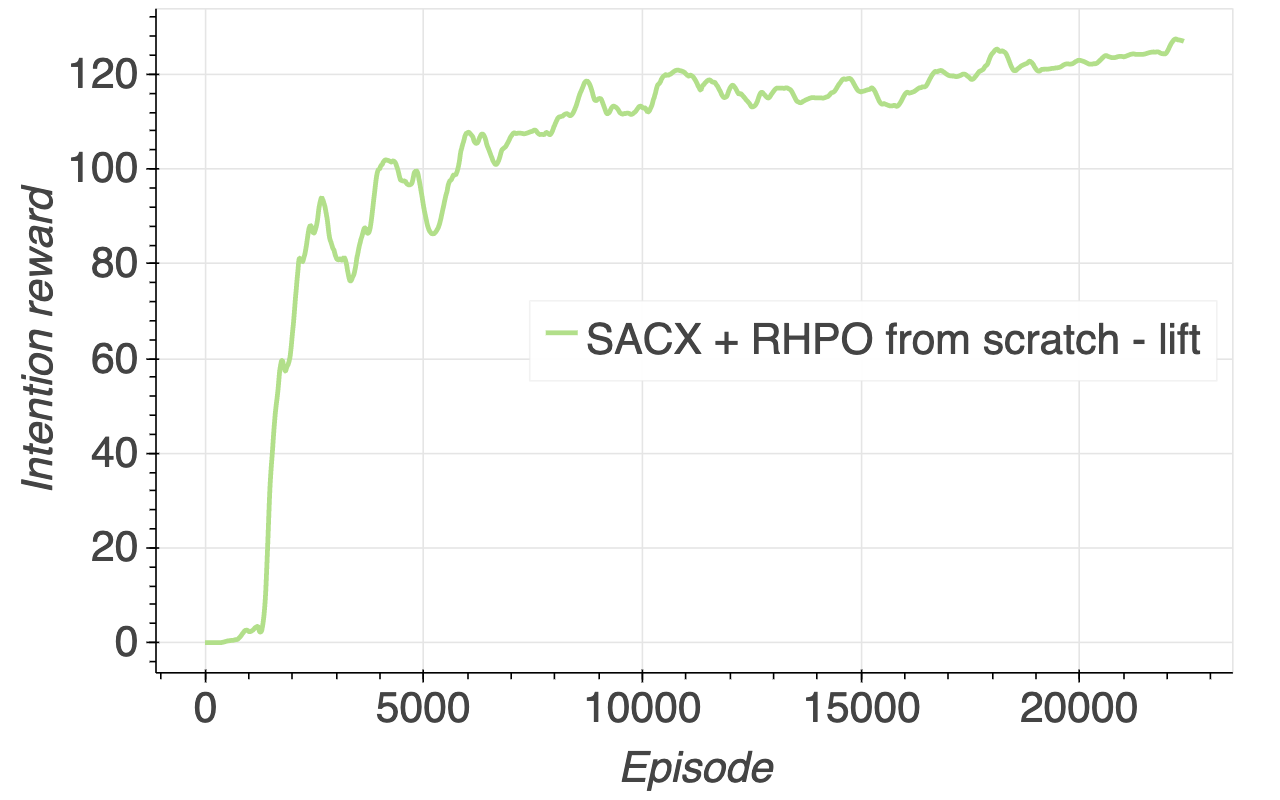}
      \caption{}
      \label{fig:3_scratch}
    \end{subfigure}
    
    \begin{subfigure}{.3\textwidth}
     \centering
      \includegraphics[width=1.\linewidth]{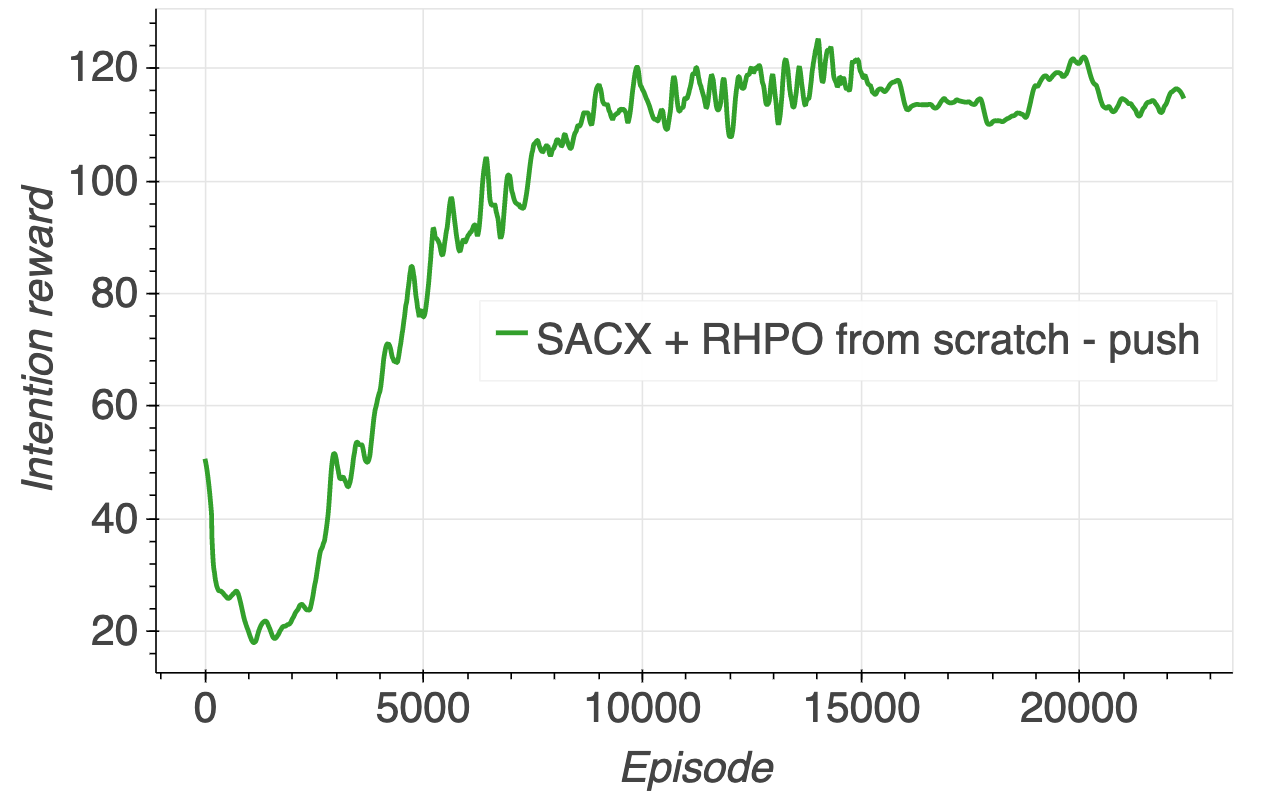}
      \caption{}
      \label{fig:4_scratch}
    \end{subfigure}
    \begin{subfigure}{.3\textwidth}
     \centering
      \includegraphics[width=1.\linewidth]{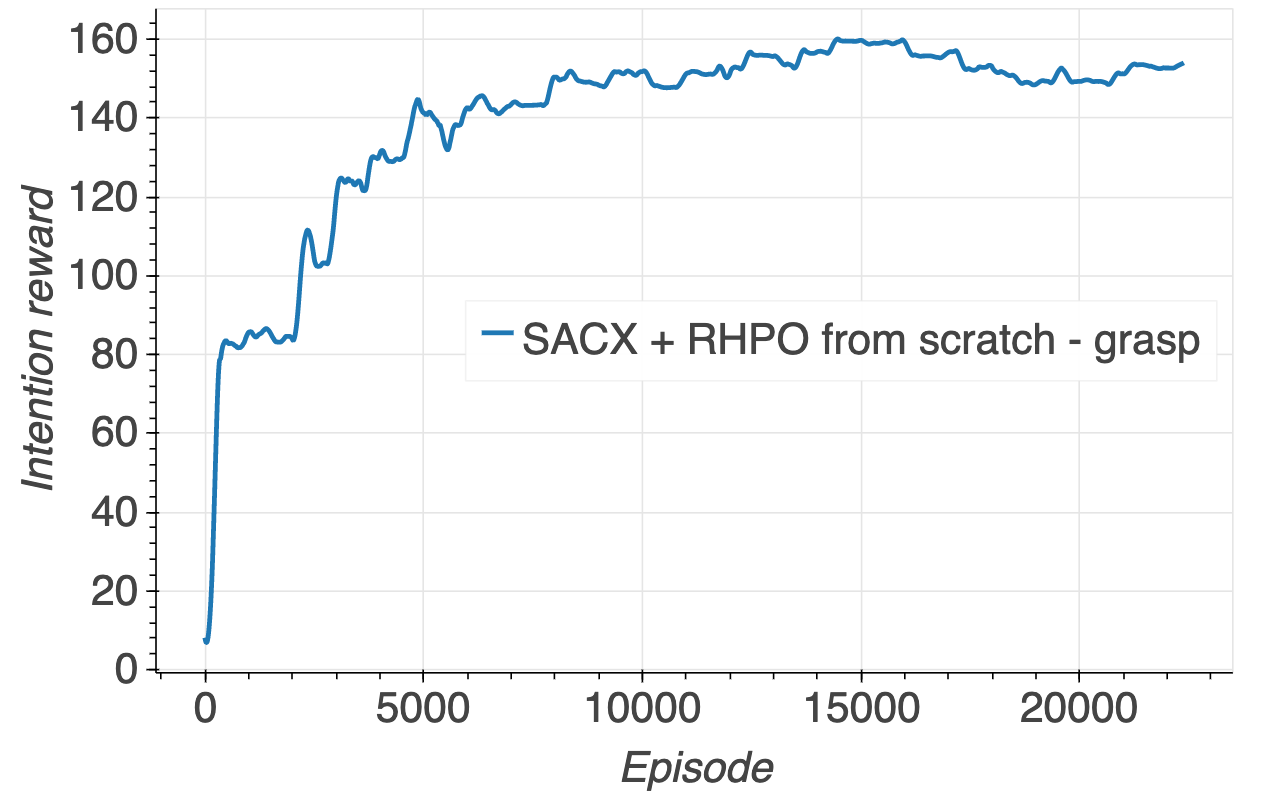}
      \caption{}
      \label{fig:5_scratch}
    \end{subfigure}
    \begin{subfigure}{.3\textwidth}
     \centering
      \includegraphics[width=1.\linewidth]{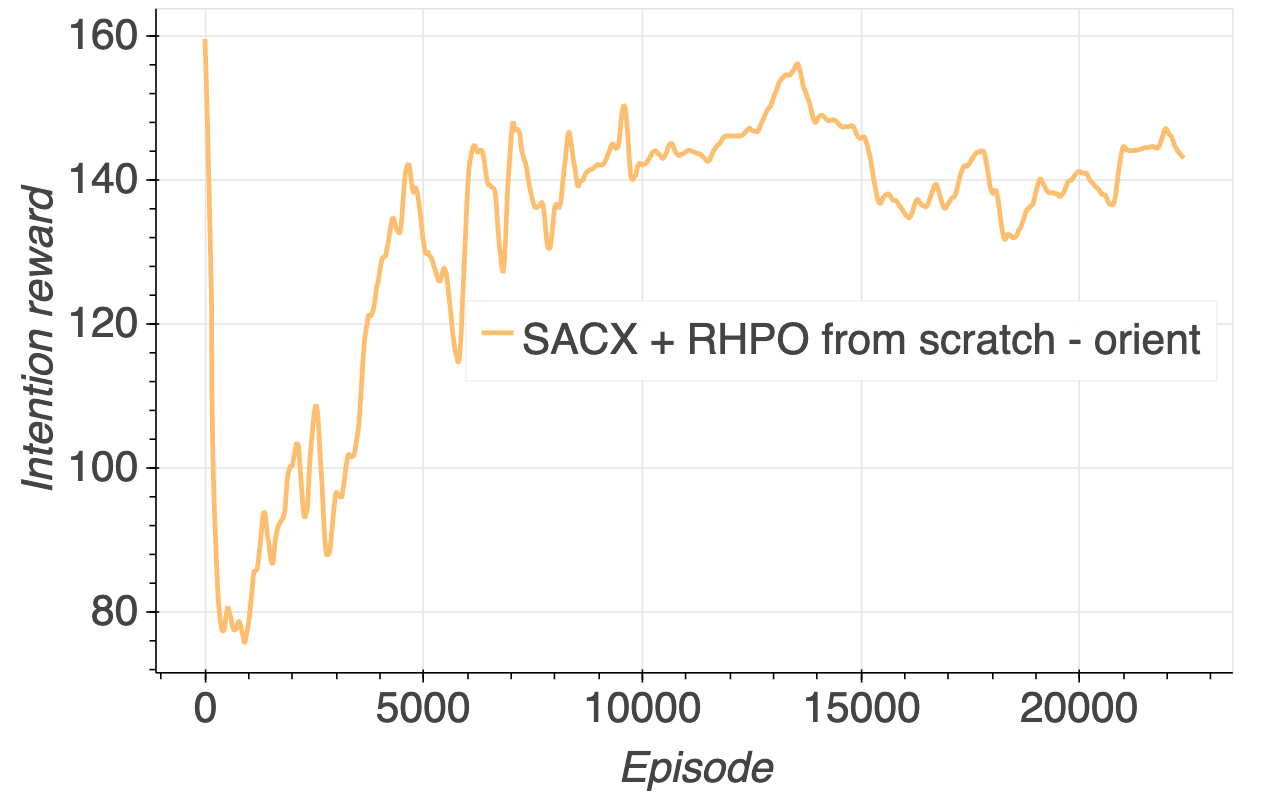}
      \caption{}
      \label{fig:6_scratch}
    \end{subfigure}
    \begin{subfigure}{.3\textwidth}
     \centering
      \includegraphics[width=1.\linewidth]{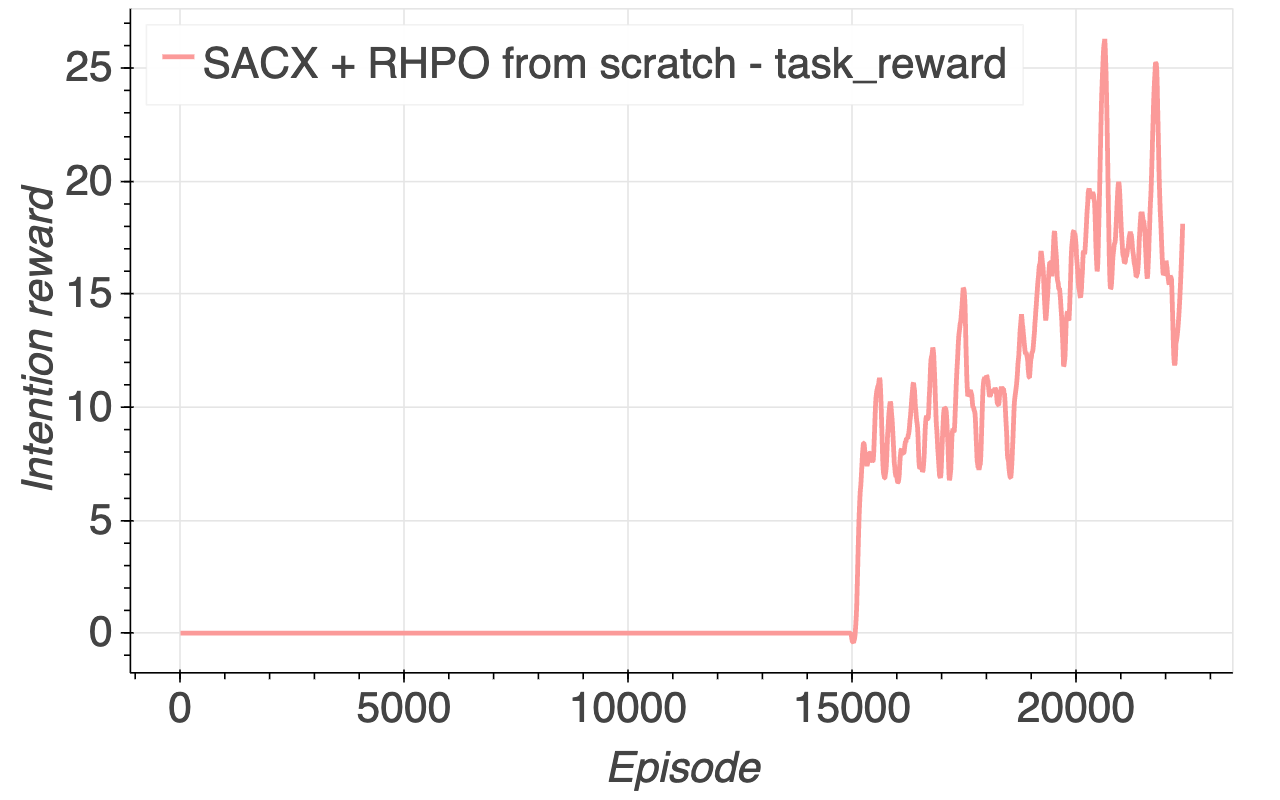}
      \caption{}
      \label{fig:7_scratch}
    \end{subfigure}
    \caption{Averaged rewards (window size: 30) in the multi-task problem formulation when training  SAC-X combined with RHPO from scratch. Figs.~\ref{fig:1_scratch} -~\ref{fig:6_scratch} show the reward for the auxiliary skills useful for solving the insertion task, while Fig.~\ref{fig:7_scratch} shows the main task reward. The plots show how the task can only be solved once the robot has learned all the useful skills.}
    \label{fig:intentions_scratch}
\end{figure}

\begin{figure}
    \centering
    \includegraphics[width=.6\linewidth]{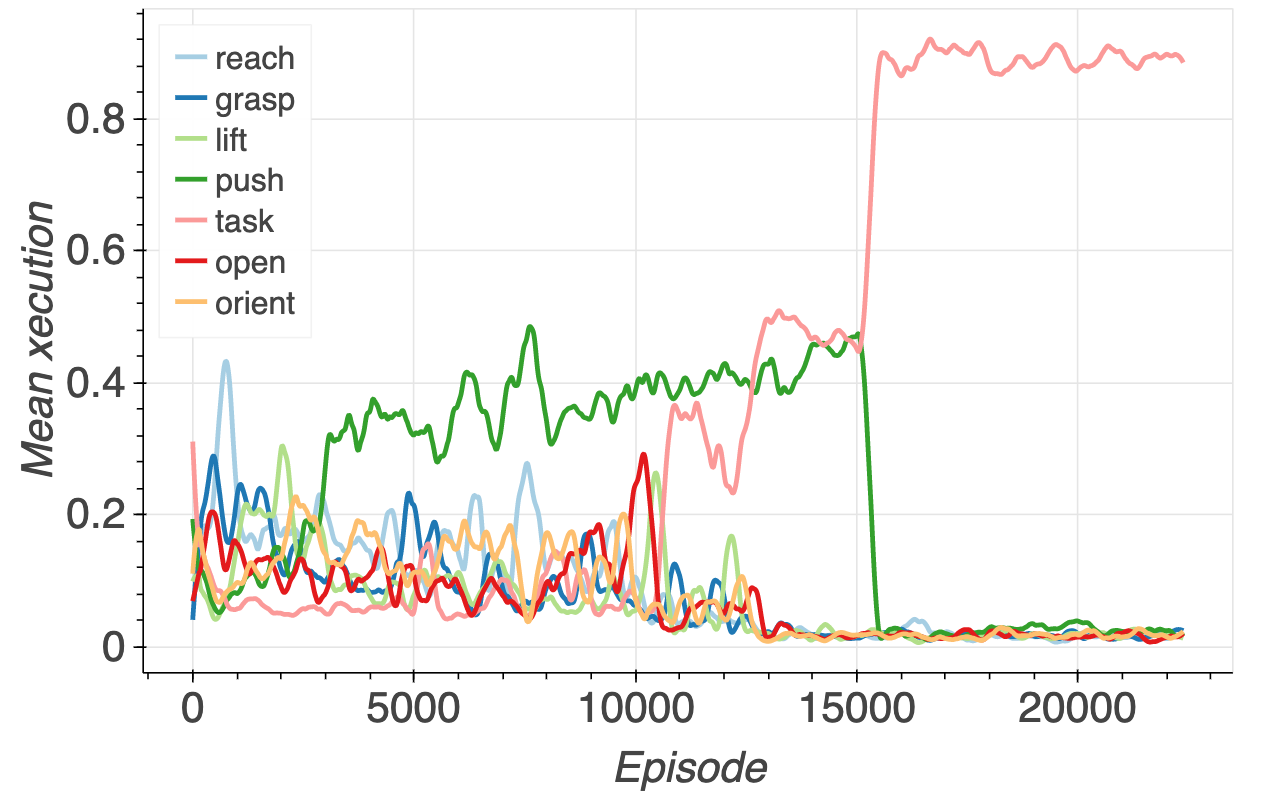}
    \caption{Mean intention execution during the training (SAC-X combined with RHPO from scratch). While the scheduler at the beginning chooses randomly among all the intentions, once the task is properly learned (after 15000 episodes), the task intention becomes the most executed intention.}
    \label{fig:scheduler_scratch}
\end{figure}

Fig.~\ref{fig:intentions_scratch} shows the rewards achieved when training the main task and the skills using SAC-X combined with RHPO from scratch on the real robot. Fig.~\ref{fig:scheduler_scratch} shows instead how the scheduler switches the execution of the intentions and the task during the training. Once the insertion task is properly learned (after 15000 episodes), the scheduler favors the execution of the main task intention.

\subsection{SAC-X combined with RHPO - with new and reloaded data}
\begin{figure}
    \centering
    \begin{subfigure}{.3\textwidth}
        \centering
      \includegraphics[width=1.\linewidth]{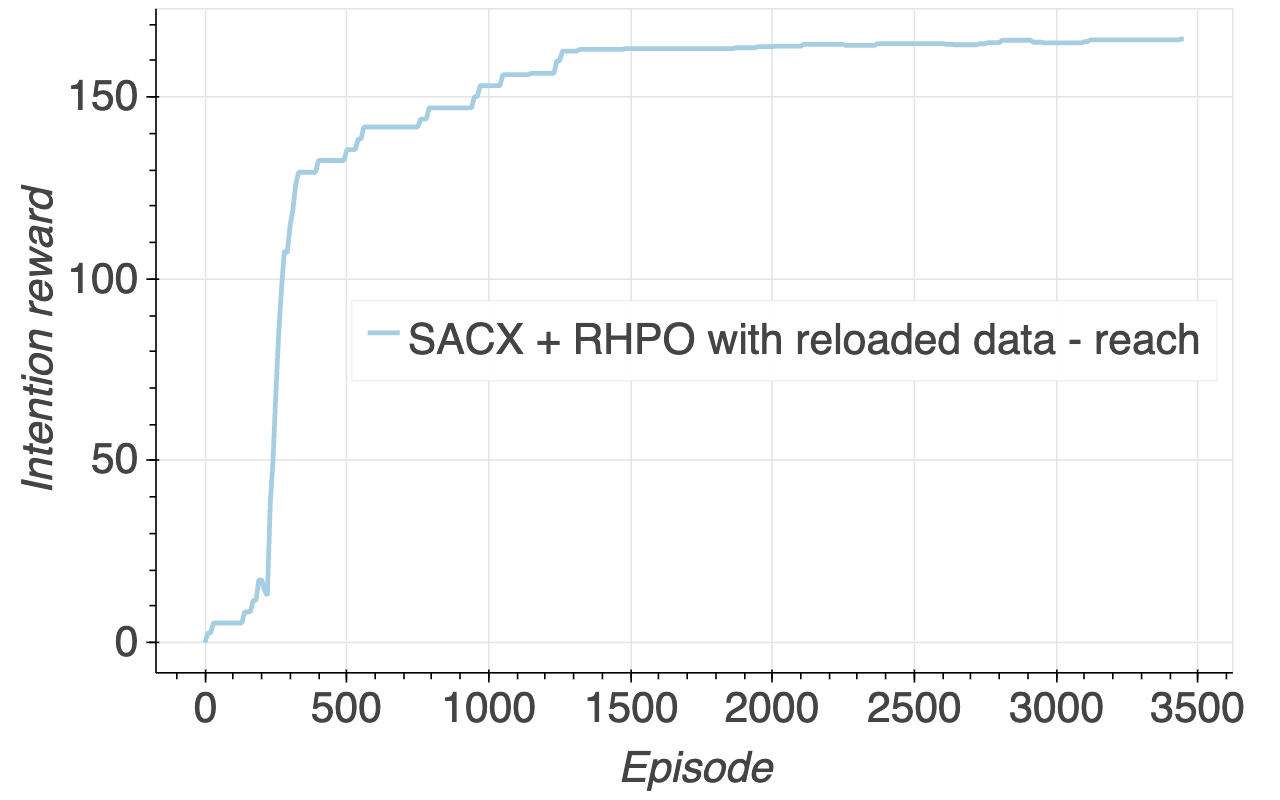}
      \caption{}
      \label{fig:1}
    \end{subfigure}
     \begin{subfigure}{.3\textwidth}
     \centering
      \includegraphics[width=1.\linewidth]{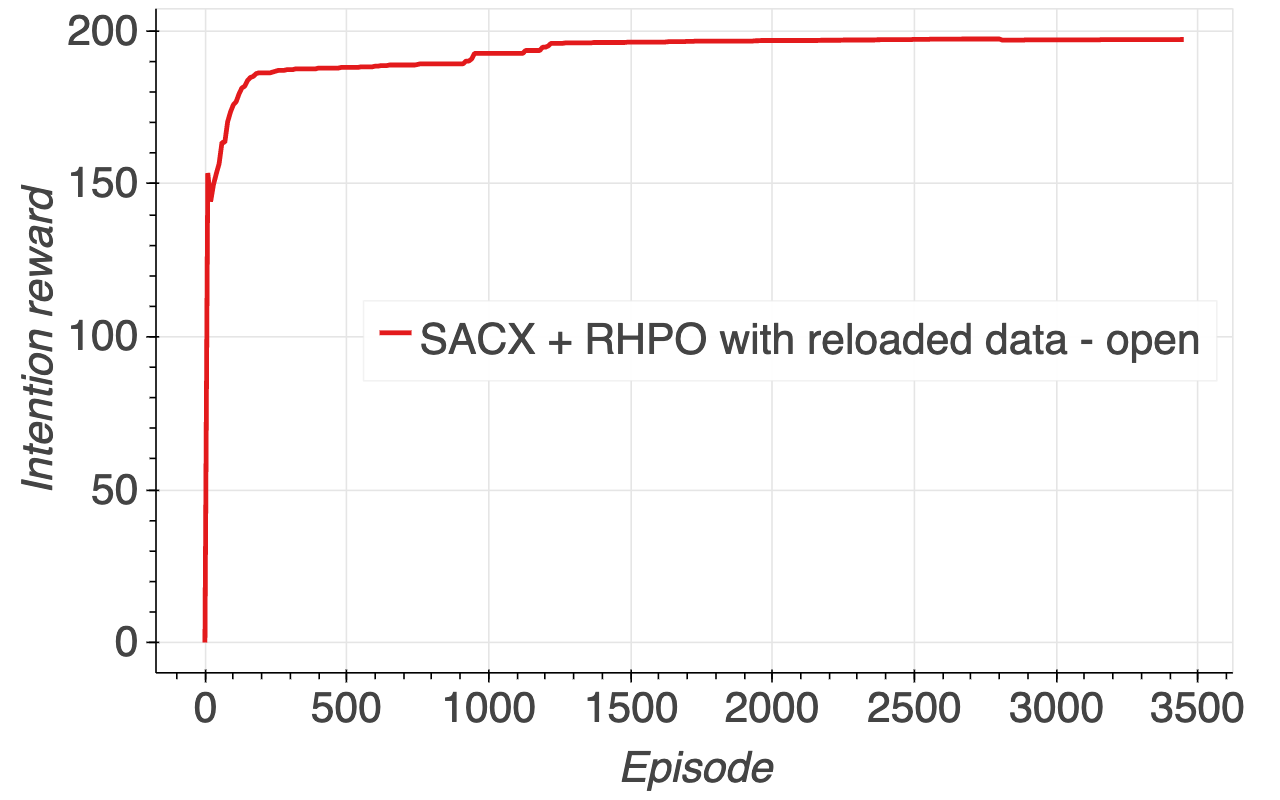}
      \caption{}
      \label{fig:2}
    \end{subfigure}
    \begin{subfigure}{.3\textwidth}
     \centering
      \includegraphics[width=1.\linewidth]{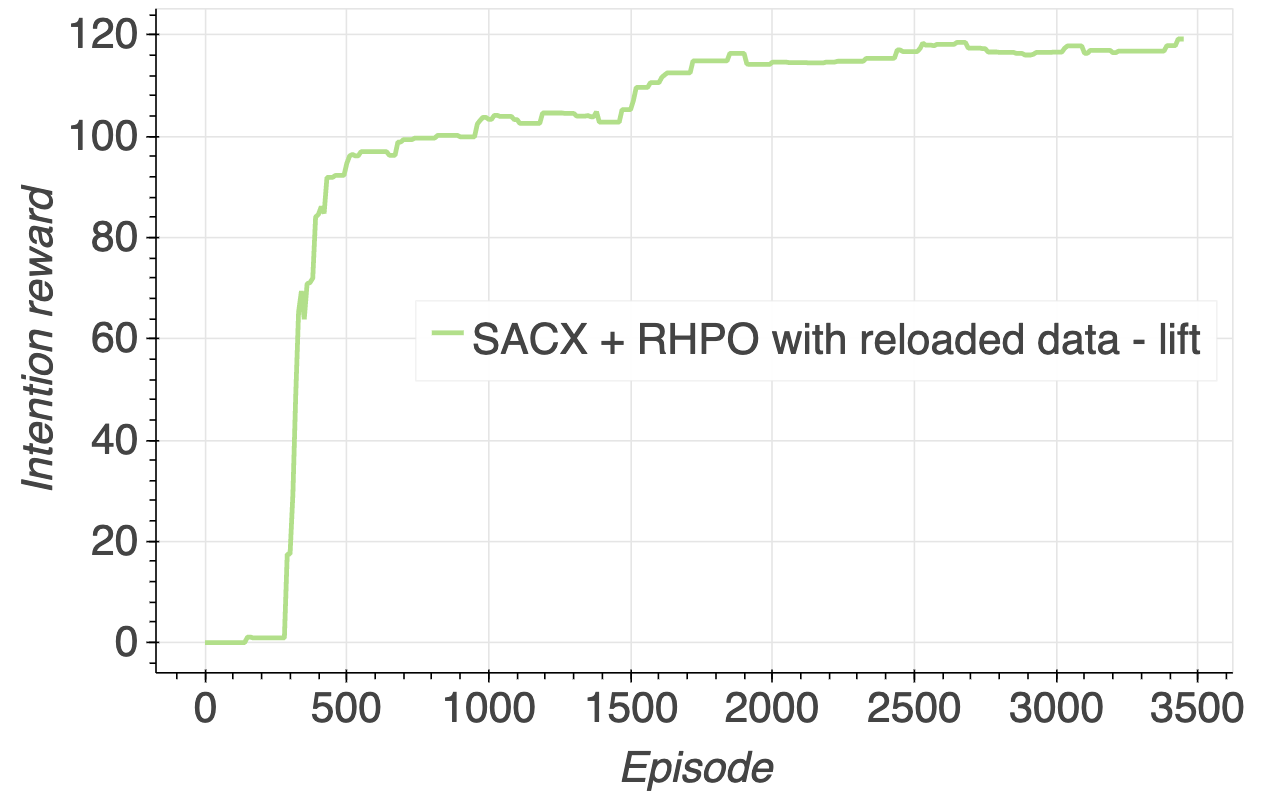}
      \caption{}
      \label{fig:3}
    \end{subfigure}
    
    \begin{subfigure}{.3\textwidth}
     \centering
      \includegraphics[width=1.\linewidth]{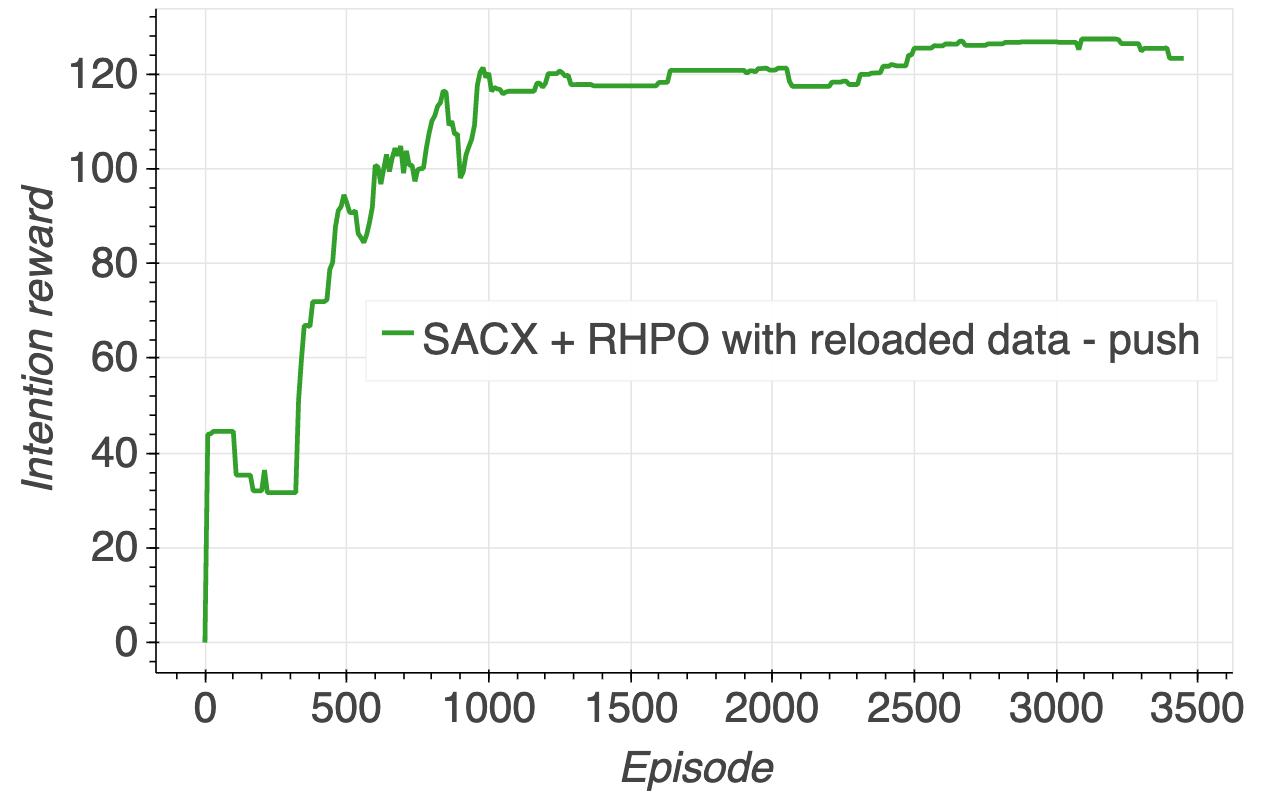}
      \caption{}
      \label{fig:4}
    \end{subfigure}
    \begin{subfigure}{.3\textwidth}
     \centering
      \includegraphics[width=1.\linewidth]{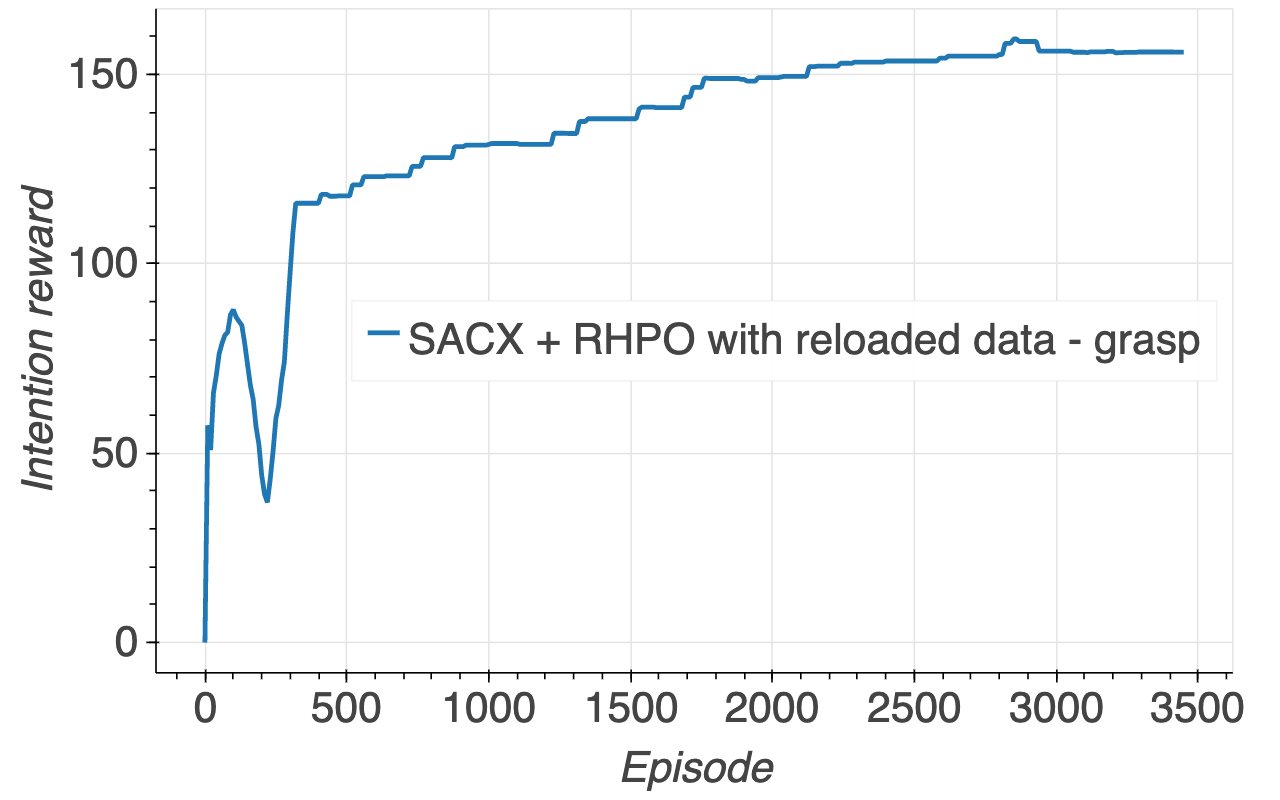}
      \caption{}
      \label{fig:5}
    \end{subfigure}
    \begin{subfigure}{.3\textwidth}
     \centering
      \includegraphics[width=1.\linewidth]{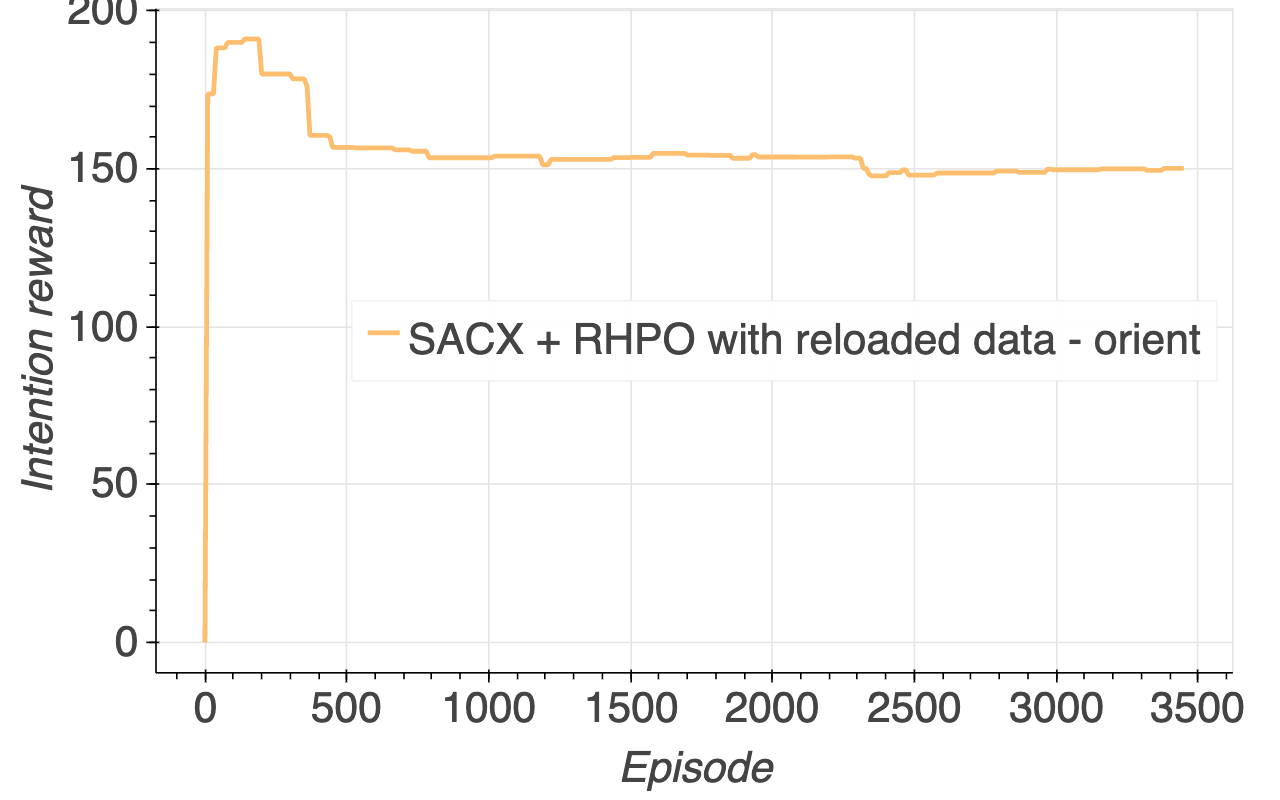}
      \caption{}
      \label{fig:6}
    \end{subfigure}
    \begin{subfigure}{.3\textwidth}
     \centering
      \includegraphics[width=1.\linewidth]{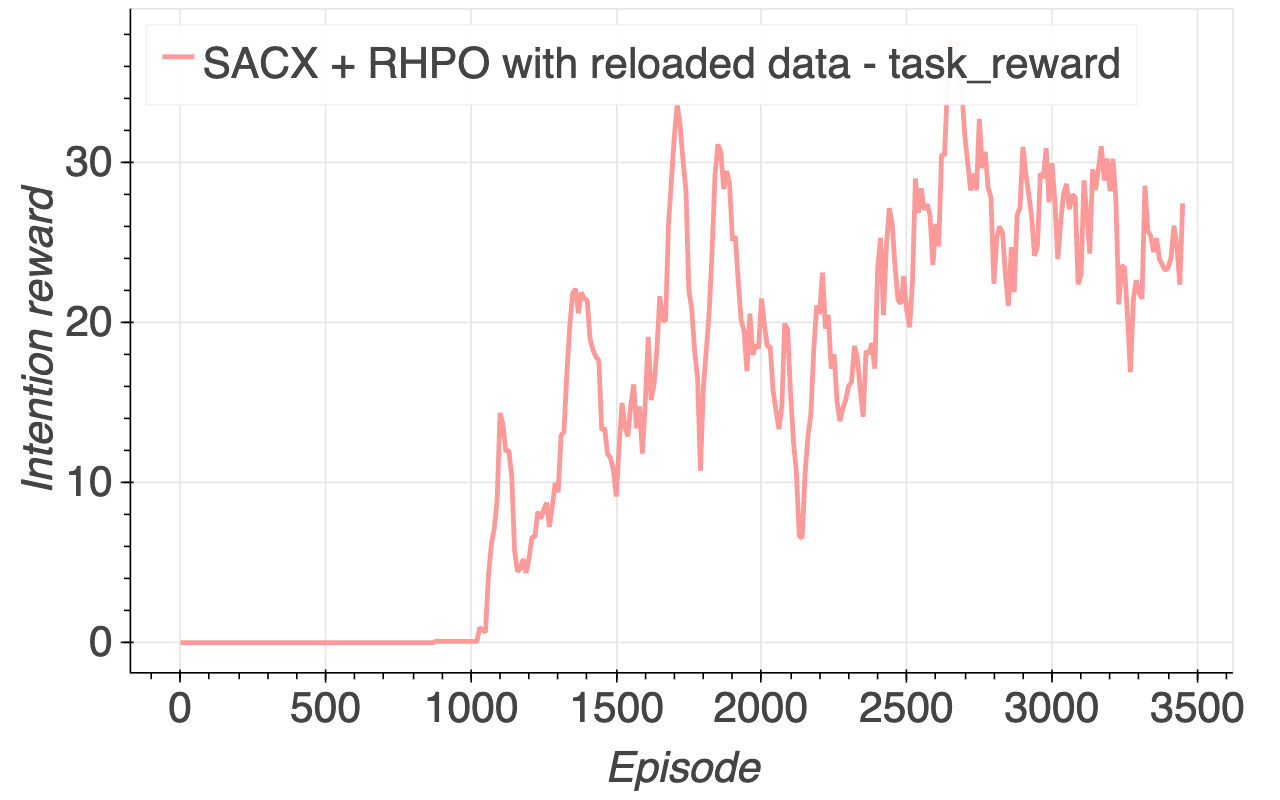}
      \caption{}
      \label{fig:7}
    \end{subfigure}
    \caption{Averaged rewards (window size: 30) in the multi-task problem formulation when using SAC-X combined with RHPO and reloading the data from the experiment from scratch. Figs.~\ref{fig:1} -~\ref{fig:6} show the reward for the auxiliary skills useful for solving the insertion task, while Fig.~\ref{fig:7} shows the task reward. The plots show how the task can only be solved once the robot has learned the useful skills.}
    \label{fig:intentions_reloaded}
\end{figure}

\begin{figure}
    \centering
    \includegraphics[width=.6\linewidth]{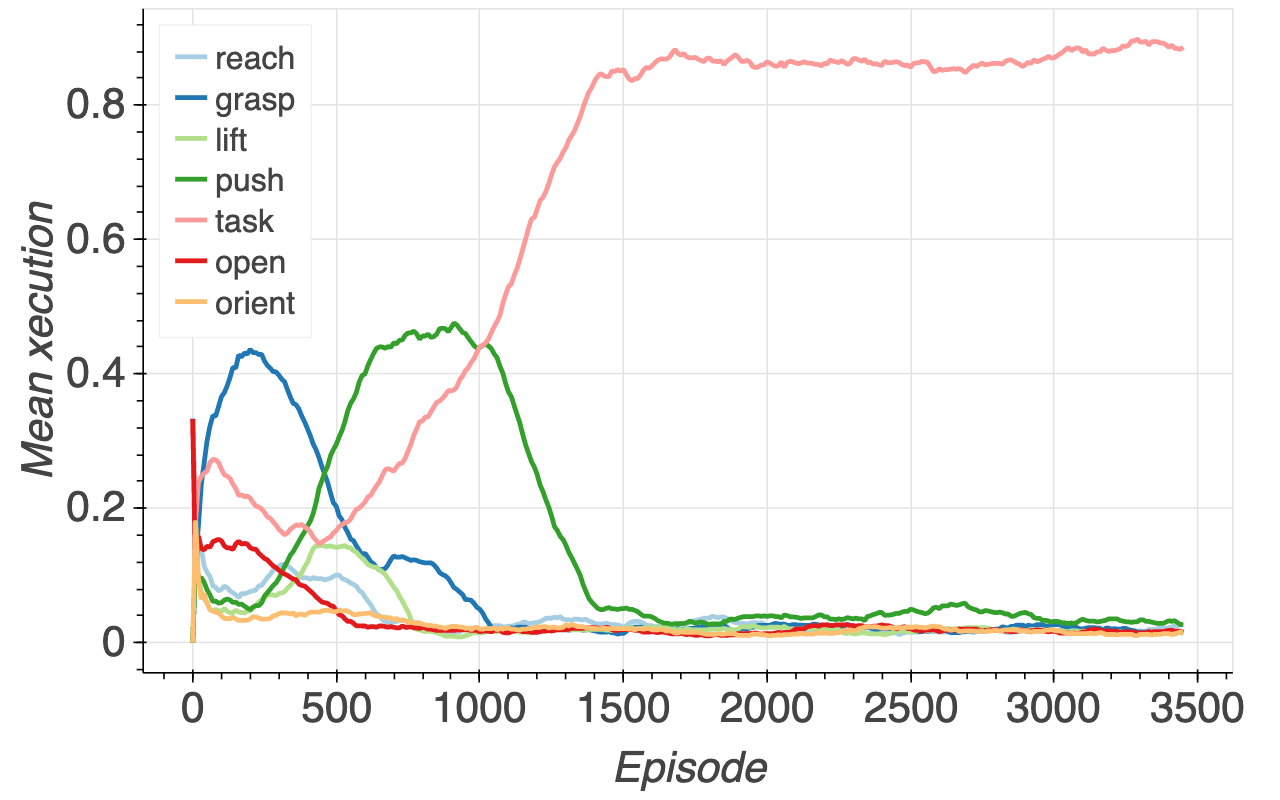}
    \caption{Mean intention execution during the training with reloaded data. While the scheduler at the beginning chooses randomly among all the intentions, once the task is learned (after 1000 episodes) the task intention becomes the most executed intention.}
    \label{fig:scheduler_reloaded}
\end{figure}

As in the previous section, Fig.~\ref{fig:intentions_reloaded} shows the rewards achieved when training on the real robot the main task and the skills using SAC-X combined with RHPO and reloading the data from the experiment from scratch. Fig.~\ref{fig:scheduler_reloaded} shows instead how the scheduler switches the intentions during the training.

\end{document}